\pdfoutput=1
\documentclass[11pt]{article}

\usepackage[final]{acl} 
\usepackage{graphicx} 
\usepackage{amsmath}
\usepackage{changepage}
\usepackage{stix,bbding,pifont,utfsym,fontawesome}
\usepackage{adjustbox}
\usepackage{array}
\usepackage{arydshln}
\usepackage{hyperref}
\usepackage{tabu}
\usepackage{amsfonts}
\usepackage{longtable}
\usepackage{booktabs}
\usepackage{xtab}
\usepackage{xltabular}
\usepackage{makecell}
\usepackage{multirow}

\title{\textrm{{\small{small}} Models, {\huge \textsc{BIG}} Impact: \\ Efficient Corpus and Graph-Based Adaptation of Small Multilingual Language Models for Low-Resource Languages}}

\author{Daniil Gurgurov\textsuperscript{\normalfont1,3} \quad Ivan Vykopal\textsuperscript{\normalfont2,4} \quad Josef van Genabith\textsuperscript{\normalfont3} \quad Simon Ostermann\textsuperscript{\normalfont3}\\
    \textsuperscript{1}University of Saarland\\ 
    \textsuperscript{2}Brno University of Technology\\
    \textsuperscript{3}German Research Center for Artificial Intelligence (DFKI)\\
    \textsuperscript{4}Kempelen Institute of Intelligent Technologies (KInIT)\\
    {\small \texttt{ \{daniil.gurgurov, josef.van\_genabith, simon.ostermann\}@dfki.de, ivan.vykopal@kinit.sk}}
    }

\begin{document}

\maketitle

\begin{abstract}

Low-resource languages (LRLs) face significant challenges in natural language processing (NLP) due to limited data. While current state-of-the-art large language models (LLMs) still struggle with LRLs, smaller multilingual models (mLMs) such as mBERT and XLM-R offer greater promise due to a better fit of their capacity to low training data sizes. This study systematically investigates parameter-efficient adapter-based methods for adapting mLMs to LRLs, evaluating three architectures: Sequential Bottleneck, Invertible Bottleneck, and Low-Rank Adaptation. Using unstructured text from GlotCC and structured knowledge from ConceptNet, we show that small adaptation datasets (e.g., up to 1 GB of free-text or a few MB of knowledge graph data) yield gains in intrinsic (masked language modeling) and extrinsic tasks (topic classification, sentiment analysis, and named entity recognition). We find that Sequential Bottleneck adapters excel in language modeling, while Invertible Bottleneck adapters slightly outperform other methods on downstream tasks due to better embedding alignment and larger parameter counts. Adapter-based methods match or outperform full fine-tuning while using far fewer parameters, and smaller mLMs prove more effective for LRLs than massive LLMs like LLaMA-3, GPT-4, and DeepSeek-R1-based distilled models. While adaptation improves performance, pre-training data size remains the dominant factor, especially for languages with extensive pre-training coverage.
The code for our experiments is available on GitHub\footnote{The code is available at \url{https://github.com/d-gurgurov/Knowledge-Driven-Adaptation-LLMs}}.
\end{abstract}

\section{Introduction} 

The need for effective natural language processing (NLP) tools for low-resource languages (LRLs) is pressing, as these languages lack sufficient data to train robust models \cite{joshi2020state, bird2022local, huang2023languagescreatedequalllms}. While \textbf{massive state-of-the-art (SoTA) large language models (LLMs)} such as GPT-4 \cite{openai2024gpt4technicalreport}, LLaMA-2 \cite{llama2}, Gemini \cite{team2023gemini}, BLOOM \cite{le2023bloom}, and the DeepSeek model family \cite{deepseekai2025deepseekr1incentivizingreasoningcapability} have demonstrated strong generalization capabilities across diverse tasks \cite{srivastava2022beyond, smith2022using, bang2023multitask}, they struggle to generalize effectively to LRLs \cite{cahyawijaya-etal-2023-nusawrites, robinson2023chatgptmtcompetitivehigh, hasan2024largelanguagemodelsspeak, adelani-etal-2024-sib}. \textbf{Smaller multilingual language models (mLMs)} like mBERT \cite{devlin2018bert} and XLM-R \cite{conneau2019unsupervised} often show greater promise for LRLs \cite{hu2020xtreme, asai2023buffet, adelani-etal-2024-comparing}.


This work investigates parameter-efficient adaptation techniques \cite{houlsby2019parameter} as an alternative to full fine-tuning, or continued pre-training, for adapting small mLMs to LRLs. We compare these approaches with the zero- and few-shot prompting and adapter-based adaptation of LLMs. Following \citet{pfeiffer2020mad}, \citet{, parović2023crosslingualtransfertargetlanguageready}, and \citet{gurgurov2024adapting}, we integrate unstructured textual data and structured knowledge from knowledge graphs (KGs), exploring their complementary benefits. KGs, which encode cross-lingual semantic relationships, have been shown to be effective for various NLP tasks \cite{peters-etal-2019-knowledge, zhang-etal-2019-ernie, wang2020k}, yet remain underexplored for LRLs. On the other hand, unstructured text provides rich contextual information and is widely used for adaptation \cite{neubig2018rapidadaptationneuralmachine, han2019unsuperviseddomainadaptationcontextualized}.

Our contributions are threefold:
\begin{itemize}
    \item First, we show that \textbf{limited adaptation data yields significant gains}—up to 1 GB of free text or a few MB of KG data. We evaluate three adapter architectures: Sequential Bottleneck, Invertible Bottleneck, and Low-Rank Adaptation \cite{houlsby2019parameter, pfeiffer2020mad, hou2022adapters}. Sequential Bottleneck excels in language modeling, while Invertible Bottleneck outperforms others on downstream tasks, likely due to differing parameterization. Adapter-based approaches match or outperform full fine-tuning while using fewer trainable parameters.  
    \item Second, \textbf{we highlight the effectiveness of smaller mLMs, such as XLM-R, for LRLs, outperforming both few-shot prompting and adaptation of massive SoTA LLMs} such as GPT-3.5 \cite{ouyang2022training}, LLaMA- 3 \cite{grattafiori2024llama3herdmodels}, and DeepSeek-R1-based distilled models \cite{deepseekai2025deepseekr1incentivizingreasoningcapability}. 
    This is in line with prior work suggesting that smaller models better align cross-lingual representations under constrained capacity \cite{wu2019emerging, dufter2020identifying, yong2023bloom1addinglanguagesupport} and shows that small LMs are often better suited for LRLs.
    \item Finally, analyzing 30 LRLs, we \textbf{show a direct relationship between pre-training and adaptation data size and performance}, with adaptation data providing diminishing returns for languages with larger pre-training data coverage. We also observe a moderate correlation between language modeling and downstream task performance, suggesting pseudo-perplexity as a useful proxy for evaluating adaptation quality.
\end{itemize}

\section{Related Work}
To improve multilingual models for LRLs without monolingual pre-training, researchers have explored full fine-tuning, adapter-based approaches, and other auxiliary methods.  

\subsection{Full Fine-Tuning Adaptation}  
Full fine-tuning has been widely used to enhance LRL performance. \citet{neubig2018rapidadaptationneuralmachine} utilized similar-language post-training to reduce overfitting. Domain-adaptive fine-tuning \cite{han2019unsuperviseddomainadaptationcontextualized} improved contextualized models like mBERT on specific domains (e.g. Middle English). Further, language-specific fine-tuning on monolingual corpora \cite{gururangan2020don, Chau_2020} and adaptation with transliterated data \cite{muller-etal-2021-unseen} boosted performance on diverse tasks, such as dependency parsing and tagging. \citet{ebrahimi2021adaptpretrainedmultilingualmodel} showed that fine-tuning on Bible corpora improved tagging and named entity recognition in languages unseen during pre-training.  


\subsection{Adapter-Based Adaptation}  
Adapters are parameter-efficient small modules that are inserted into model layers, avoiding catastrophic forgetting \cite{french1999catastrophic}, reducing computational costs \cite{houlsby2019parameter, strubell-etal-2019-energy}, and requiring fewer training examples \cite{faisal2022phylogenyinspiredadaptationmultilingualmodels}. Frameworks like MAD-X \cite{pfeiffer2020mad} introduced language and task adapters, improving named entity recognition. Extensions such as UDapter \cite{udapterlanguageadaptationtruly} and MAD-G \cite{ansell-etal-2021-mad-g} leveraged typological features for improved zero-shot inference. Hierarchical adapters based on language phylogeny \cite{faisal2022phylogenyinspiredadaptationmultilingualmodels}, methods addressing resource imbalances with language combination \cite{lee-etal-2022-fad, parovic-etal-2022-bad}, and exposing task adapters to target languages during training to address training-inference mismatches \cite{parović2023crosslingualtransfertargetlanguageready} have further advanced adapter effectiveness. Recent work \cite{pfeiffer-etal-2022-lifting, yong2023bloom1addinglanguagesupport} emphasized the efficiency of adapter-based tuning over continued pre-training for LRLs, with performance tied to data quantity.  

\subsection{Knowledge Graph Integration}  
KGs improve the quality of static word embeddings \cite{faruqui2014retrofitting, speer2017conceptnet, gurgurov2024gremlinrepositorygreenbaseline} and, more recently, LMs by leveraging structured semantic relationships, predominantly for high-resoure languages \cite{miller1995wordnet, navigli2012babelnet, speer2017conceptnet}. Approaches like KnowBERT \cite{peters-etal-2019-knowledge} and ERNIE \cite{zhang-etal-2019-ernie} improve LMs through entity linkers and attention. LIBERT \cite{lauscher2020specializingunsupervisedpretrainingmodels} incorporates semantic constraints for better task performance. CN-ADAPT \cite{lauscher2020common} and K-Adapter \cite{wang2020k} use bottleneck adapters \cite{houlsby2019parameter} to inject structured knowledge into models, improving commonsense reasoning and relational tasks. 

\section{Methodology}  
This section describes our approaches to adapting mLMs for LRLs and the data resources used.  

\subsection{Model Adaptation}  
We adapt mBERT \cite{devlin2018bert} and XLM-R-base \cite{conneau2019unsupervised} using three adapter architectures: Sequential Bottleneck (\texttt{Seq\_bn}; \citet{houlsby2019parameter, pfeiffer2020mad}), Sequential Bottleneck with Invertible Layers (\texttt{Seq\_bn\_inv}; \citet{pfeiffer2020mad}), and Low-Rank Adaptation (\texttt{LoRA}; \citet{hou2022adapters}). Additionally, we adapt LLaMA-3-8B \cite{grattafiori2024llama3herdmodels}, but exclusively with \texttt{Seq\_bn\_inv} adapters (due to computational constraints). Language adapters are pre-trained with a masked language modeling (MLM) objective \cite{devlin2018bert} for mBERT and XLM-R on structured data (ConceptNet; \citet{speer2017conceptnet}) and unstructured data (GlotCC; \citet{kargaran2024glotcc}).\footnote{Full fine-tuning is performed only on the GlotCC data for mBERT and XLM-R due to ConceptNet's limited size.} Further, we pre-train language adapters for LLaMA-3 with a causal language modeling (CLM) objective \cite{radford2018improving}, only with unstructured data, leaving the exploration of graph knowledge injection into large-scale LMs for future work. 


Task-specific adapters are trained on target language data using the \texttt{Seq\_bn} architecture. These adapters are stacked on "frozen" LMs and language adapters, following prior work \cite{pfeiffer2020mad, lee-etal-2022-fad, parović2023crosslingualtransfertargetlanguageready}. We also experiment with adapter fusion \cite{pfeiffer2020adapterfusion}, combining language adapters trained on different data types.  

\subsection{Data Sources}  
\paragraph{Structured Data.} ConceptNet \cite{speer2017conceptnet}, a multilingual knowledge graph, provides common-sense knowledge across 304 languages. We preprocess the data by converting ConceptNet triples into natural language sentences, similar to \citet{lauscher2020common} and \citet{gurgurov2024adapting}, using predefined predicates (Appendix \ref{app:cn_conv}), and split it into train and validation sets.  

\paragraph{Unstructured Data.} GlotCC-V1 \cite{kargaran2024glotcc} is a large-scale multilingual corpus derived from CommonCrawl \cite{wenzek-etal-2020-ccnet}. It emphasizes LRLs, providing high-quality text in 1,000 languages. To simulate a low-resource environment for all languages, we limit each language to 1 GB (if it exceeds this limit), clean the data, and split it into training and validation sets. 

\section{Experimental Setup}  
This section details the experimental setup, including language selection, evaluation tasks, and adapter training procedures.  

\subsection{Languages}  
We selected 30 diverse LRLs identified by \citet{joshi2020state} as low-resource—representing a diverse set that includes \textit{Thai, Romanian, Bulgarian, Danish, Greek, Hebrew, Slovak, Slovenian, Latvian, Indonesian, Georgian, Bengali, Azerbaijani, Urdu, Macedonian, Telugu, Nepali, Marathi, Swahili, Welsh, Uzbek, Javanese, Sundanese, Sinhala, Amharic, Kurdish, Uyghur, Maltese, Tibetan, and Yoruba}—to evaluate adapter performance across underrepresented linguistic contexts. Table \ref{tab:language-stats} (Appendix \ref{app:lang_det}) summarizes pre-training and adaptation data availability, inclusion in mLM pre-training, and other language-specific details.

\begin{table*}[ht!]
    \centering
    \begin{adjustbox}{max width=\textwidth}
    \begin{tabular}{llcccccccc}
        \toprule
        \multirow{2}{*}{\textbf{Model}} & \multirow{2}{*}{\textbf{Configuration}} & \multicolumn{2}{c}{\textbf{TC} ($\uparrow$)} & \multicolumn{2}{c}{\textbf{NER} ($\uparrow$)} & \multicolumn{2}{c}{\textbf{SA} ($\uparrow$)} & \multicolumn{2}{c}{\textbf{MLM} ($\downarrow$)} \\
        \cmidrule(lr){3-4} \cmidrule(lr){5-6} \cmidrule(lr){7-8} \cmidrule(lr){9-10}
        & & \textbf{Seen} & \textbf{Unseen} & \textbf{Seen} & \textbf{Unseen} & \textbf{Seen} & \textbf{Unseen} & \textbf{Seen} & \textbf{Unseen} \\
        \midrule
        \multirow{10}{*}{mBERT} & Baseline & 77.67 & 28.72 & 83.82 & 42.54 & 82.18 & 71.03 & 25.17 & \textbf{124.67} \\
        & + \texttt{LoRA} (Glot) & 78.74 & 36.65 & 84.2 & 44.51 & 82.75 & 73.27 & 10.44 & 7434.61 \\
        & + \texttt{Seq\_bn} (Glot) & 79.28 & 41.42 & 84.46 & 45.04 & 82.99 & 73.3 & \textbf{8.95} & 12218.65 \\
        & + \texttt{Seq\_bn\_inv} (Glot) & \textbf{79.35} & \textbf{42.4} & 84.36 & \textbf{45.64} & \textbf{83.64} & \textbf{73.91} & 14.31 & 27170.23 \\
        & + \texttt{LoRA} (ConceptNet) & 77.87 & 24.88 & 84.38 & 41.32 & 82.59 & 70.79 & 37.37 & 126.44 \\
        & + \texttt{Seq\_bn} (ConceptNet) & 78.39 & 25.87 & 84.35 & 41.2 & 81.9 & 70.48 & 41.22 & 139.25 \\
        & + \texttt{Seq\_bn\_inv} (ConceptNet) & 78.42 & 24.18 & \textbf{84.7} & 41.48 & 81.58 & 71.54 & 55.95 & 157.49 \\
        & + \texttt{Seq\_bn} (Glot+ConceptNet) & -- & -- & 84.36 & 44.21 & -- & -- & -- & -- \\
        & + \texttt{Seq\_bn\_inv} (Glot+ConceptNet) & -- & -- & 84.36 & 44.93 & -- & -- & -- & -- \\
        \hdashline
        & Full Fine-tune & \underline{81.73} & \underline{43.65} & -- & -- & \underline{84.07} & \underline{73.97} & 9.25 & 81492.4 \\
        \midrule
        \multirow{10}{*}{XLM-R} & Baseline & 81.14 & 34.52 & 77.33 & 54.45 & 87.45 & 60.72 & 15.65 & 203.96 \\
        & + \texttt{LoRA} (Glot) & 82.31 & 40.94 & 77.52 & 52.01 & 87.98 & 62.02 & 6.83 & \textbf{97.99} \\
        & + \texttt{Seq\_bn} (Glot) & 83.63 & 49.72 & 78.57 & 54.4 & \textbf{88.2} & \textbf{65.94} & \textbf{6.53} & 122.08 \\
        & + \texttt{Seq\_bn\_inv} (Glot) & \textbf{84.06} & \textbf{51.43} & 78.17 & 55.64 & \textbf{88.2} & 65.88 & 10.56 & 713.65 \\
        & + \texttt{LoRA} (ConceptNet) & 80.71 & 29.08 & 78.38 & 52.71 & 87.48 & 60.00 & 20.29 & 902.31 \\
        & + \texttt{Seq\_bn} (ConceptNet) & 80.82 & 33.19 & 77.64 & 49.39 & 87.09 & 58.64 & 20.01 & 482.22 \\
        & + \texttt{Seq\_bn\_inv} (ConceptNet) & 80.64 & 33.59 & 78.62 & 51.04 & 87.28 & 59.52 & 22.81 & 569.48 \\
        & + \texttt{Seq\_bn} (Glot+ConceptNet) & -- & -- & \textbf{80.83} & \textbf{61.83} & -- & -- & -- & -- \\
        & + \texttt{Seq\_bn\_inv} (Glot+ConceptNet) & -- & -- & 80.68 & 60.31 & -- & -- & -- & -- \\
        \hdashline
        & Full Fine-tune & \underline{85.61} & \underline{57.3} & -- & -- & \underline{88.56} & \underline{68.19} & 10.57 & 206.68 \\
        \bottomrule
    \end{tabular}
    \end{adjustbox}
    \caption{Results for mBERT and XLM-R across 4 tasks: Topic Classification (TC), Named Entity Recognition (NER), Sentiment Analysis (SA), Masked Language Modeling (MLM). All numbers are the averages for the 30 studied LRLs and provided separately for the languages included ("seen") and languages not included ("unseen") in the pre-training data of a model. The baselines are the models with a single task adapter for downstream tasks, or without adapters for MLM. The full results for each task are in the Appendix.}
    \label{tab:main_results}
\end{table*}

\subsection{Language Adapter Training}  
Language adapters were trained on mBERT and XLM-R for all languages using MLM with GlotCC and ConceptNet data. We evaluated \texttt{Seq\_bn}, \texttt{Seq\_bn\_inv}, and \texttt{LoRA}, with the default hyperparameters (Appendix~\ref{app:adapt_params}). For LLaMA-3-8B, only GlotCC data was used with the \texttt{Seq\_bn\_inv} architecture and CLM objective for a subset of 5 languages due to computational constraints. Training consisted of up to 100,000 steps for GlotCC and 25,000 steps for ConceptNet, with a batch size of 16 and learning rate of 1e-4. 

\subsection{Task-Specific Training}  
Adapters were evaluated on four tasks. For \textit{Masked Language Modeling} (MLM), we used the FLORES-200 devtest set \cite{nllbteam2022languageleftbehindscaling}, comprising 1012 parallel sentences, and measured pseudo-perplexity \cite{salazar2019masked} as a proxy for linguistic acceptability. \textit{Topic Classification} (TC) employed the 7-class SIB-200 dataset \cite{adelani-etal-2024-sib}, training task adapters on predefined splits (701 train, 99 validation, 204 test examples) and fixed hyperparameters (Appendix~\ref{app:adapt_params}), with F1 scores computed on the test set \cite{sokolova2006beyond}. For \textit{Sentiment Analysis} (SA), binary-class datasets from multiple sources (Table \ref{tab:sa-sources} in Appendix \ref{app:sa_data}) were used to train task adapters with similar hyperparameters, evaluating performance via F1 scores. Finally, \textit{Named Entity Recognition} (NER) used the WikiANN dataset \cite{pan-etal-2017-cross}, with data distributions detailed in Table \ref{tab:ner-sources} (Appendix \ref{app:ner_data}), and was evaluated with the "seqeval" F1 score \cite{seqeval}. 

\subsection{Baselines}  
For MLM, mBERT and XLM-R were evaluated without adapters; LLaMA-3 was not evaluated on this task due to its autoregressive nature. For TC, SA, and NER, baselines used a single \texttt{Seq\_bn} task adapter, isolating the impact of language adapters and enabling direct comparisons with language adapter-enhanced models.

\section{Results: Small mLMs}
This section summarizes the outcomes of the mLM adaptation experiments. Tables \ref{tab:main_results} and \ref{tab:llms_adapter} report the average results across 30 selected LRLs.

\subsection{Masked Language Modeling}
Glot-based adapters substantially improved pseudo-perplexity (Tables \ref{tab:mbert_pppl} and \ref{tab:xlm_r_pppl} Appendices \ref{app:mlm_1} and \ref{app:mlm_2}), particularly for mBERT. The \texttt{Seq\_bn} adapter achieved the largest reduction, averaging a 65\% improvement, followed by \texttt{LoRA} and \texttt{Seq\_bn\_inv}. For XLM-R, \texttt{Seq\_bn} also excelled overall, while \texttt{LoRA} performed better for high-resource languages. In contrast, ConceptNet-based adapters did not enhance MLM performance, likely due to the dataset's limited size and structured nature, but showed utility in downstream tasks (Section \ref{sec:dt_section}).


Full fine-tuning on GlotCC generally outperformed language adapters for mBERT (Table \ref{tab:mbert_pppl}), while adapters applied to XLM-R often surpassed full fine-tuning (Table \ref{tab:xlm_r_pppl}). Compared to larger models, Glot-based XLM-R adapters outperformed Glot500-m \cite{imanigooghari-etal-2023-glot500}, despite the latter's larger vocabulary and more extensive training data. The performance of Glot500-m likely reflects its sampling strategy, which heavily prioritizes LRLs. Additionally, XLM-R-large without language adapters \cite{conneau2019unsupervised} slightly surpassed XLM-R-base with adapters (Appendix \ref{ap:adaptervsbigmodels}). 

\subsection{Downstream Tasks}
\label{sec:dt_section}
We further fine-tuned task adapters stacked on language adapters and mLMs. The detailed results are in Tables \ref{tab:model_comparison_tc_mbert}, \ref{tab:model_comparison_tc_xlm-r}, \ref{tab:ner_mbert}, \ref{tab:ner_xlm-r}, \ref{tab:sa_mbert}, and \ref{tab:sa_xlmr} (Appendices \ref{app:tc_1}, \ref{app:tc_2}, \ref{app:ner_1}, \ref{app:ner_2}, \ref{app:sa_1}, and \ref{app:sa_2}).\footnote{Below, we report the average scores across languages for each configuration. Notably, numerous individual languages show improvements under each configuration.}


\subsubsection{Topic Classification}
ConceptNet-based adapters showed marginal average improvements over the baseline. For mBERT, \texttt{Seq\_bn\_inv} primarily improved F1 scores for languages included in pre-training, but gains were inconsistent for others. Glot-based adapters demonstrated more substantial improvements, particularly for languages with less pre-training data. \texttt{Seq\_bn\_inv} achieved the best performance across both models, with mBERT showing an average 2-point F1 improvement for seen languages and 14 points for unseen ones, while XLM-R exhibited an average boost of 3 points for pre-trained languages and 17 points for excluded ones. Full fine-tuning provided better average results for both mBERT—4 points for seen and 15 points for unseen languages, and XLM-R—4 points and 23 points, respectively-with adapters being slightly behind. Additional experiments with \texttt{Seq\_bn\_inv} on LLaMA-3 showed an average 28-point improvement over single-task adapter setups.


\subsubsection{Named Entity Recognition}
For mBERT, ConceptNet adapters provided modest average improvements mostly for seen languages, with \texttt{Seq\_bn\_inv} achieving the highest gains of 1 F1 point on average. Glot-based adapters offered slightly lower gains for seen languages (0.5 points) 
but larger improvements for unseen ones, with \texttt{Seq\_bn\_inv} delivering an average gain of 3 points. XLM-R exhibited similar trends: ConceptNet adapters improved average scores by 1 point (\texttt{Seq\_bn\_inv}) for seen languages but showed decreases for unseen ones, while Glot-based adapters reached a 0.5-point improvement (\texttt{Seq\_bn\_inv}) for seen languages and 1 point for unseen ones. 
Meanwhile, LLaMA-3 with \texttt{Seq\_bn\_inv} failed to outperform its baseline.

Due to NER benefiting the most from ConceptNet adapters, we also experimented with the combination of ConceptNet and Glot adapters (\texttt{Seq\_bn} and \texttt{Seq\_bn\_inv}) with adapter fusion \cite{pfeiffer2020adapterfusion}. This provided the greatest benefits for XLM-R, boosting F1 scores by up to 3 points for seen languages and 7 points for unseen ones, outperforming both individual adapters and the baselines. For mBERT, however, fusion did not produce additional improvements.

\begin{table}[t]
\centering
\small
\resizebox{\columnwidth}{!}{
\begin{tabular}{@{}llcc@{}}
\toprule
\textbf{Model} & \textbf{\#Params (B)} & \textbf{TC} ($\uparrow$) & \textbf{NER} ($\uparrow$) \\ 
\midrule
mBERT+\texttt{Seq\_bn\_inv}  & \textbf{0.177} & \textbf{71.92} & \textbf{85.28} \\ 
XLM-R+\texttt{Seq\_bn\_inv}  & \textbf{0.279} & \textbf{80.79} & \textbf{85.42} \\
\midrule
DeepSeek-R1-D-Llama & 8 & 20.5 & - \\ 
DeepSeek-R1-D-Qwen & 14 & 41.88 & - \\ 
DeepSeek-R1-D-Qwen & 32 & 68.54 & - \\ 
DeepSeek-R1-D-Llama & 70 & 70.72 & - \\ 
\midrule
LLaMA-3 & 8 & 65.8 & - \\ 
LLaMA-3.1 & 8 & 65.62 & - \\ 
Gemma & 7 & 60.21 & - \\ 
Gemma-2  & 9 & 44.27 & - \\ 
Qwen-1.5 & 7 & 40.41 & - \\
Qwen2 & 7 & 56.82 & - \\ 
GPT-3.5-turbo-0301 & - & - & 70.65 \\ 
GPT-3.5-turbo-0613 & - & 45.02 & - \\ 
GPT-4-0613 & - & 45.82 & - \\ 
\midrule
LLaMA-2 & 7 & 18.24 & - \\ 
BLOOM & 7 & 13.02 & 31.35 \\ 
BLOOMz & 7 & 17.51 & 20.92 \\ 
mT0 & 13 & - & 17.48 \\ 
Occiglot-eu5 & 7 & 28.56 & \\ 
XGLM & 7.5 & 29.98 & - \\ 
Yayi & 7 & 16.88 & - \\ 
LLaMAX2 Alpaca & 7 & 23.13 & - \\ 
Mala-500-v2 & 10 & 5.74 & - \\ 
\bottomrule
\end{tabular}
}
\caption{Average F1 scores on overlapping LRLs for LLMs and our Glot adapter-based mLMs on TC and NER. Prompting results are 3-shot, based on \citet{ji2024emma500enhancingmassivelymultilingual} for TC and \citet{asai2023buffet} for NER. For NER, we report averages across eight overlapping languages, while the GPT-3.5 average is based on only two. TC results for GPT-3.5 and GPT-4 are zero-shot, as reported by \citet{adelani-etal-2024-sib}. DeepSeek results are zero-shot and were obtained in our evaluation. Per-language results are in Appendix \ref{ap:prompting}.}
\label{tab:llms_tc_ner}
\end{table}

\subsubsection{Sentiment Analysis}
For mBERT, ConceptNet adapters showed limited average gains, with only \texttt{LoRA} surpassing the baseline for seen languages, with a 0.25-point improvement. Glot adapters consistently performed better across all architectures, with \texttt{Seq\_bn\_inv} achieving the highest F1 scores, with a 1.5-point improvement for seen and a 3-point gain for unseen languages. For XLM-R, ConceptNet adapters exhibited no average improvements, while Glot adapters consistently enhanced performance. \texttt{Seq\_bn} and \texttt{Seq\_bn\_inv} achieved gains of up to 1 point for seen and 5 points for unseen languages. Full fine-tuning yielded similar results with a 2-point and 3-point boosts for mBERT, and 1-point and 8-point improvements respectively, for seen and unseen language groups. Finally, \texttt{Seq\_bn\_inv} on LLaMA-3 resulted in a 10-point average improvement over its baseline.



\section{Results: Small mLMs vs. SoTA LLMs}
Compared to the zero-shot prompting of proprietary LLMs like GPT-3.5-Turbo \cite{NEURIPS2022_b1efde53} and GPT-4 \cite{openai2024gpt4technicalreport} on the SIB-200 TC task \cite{adelani-etal-2024-sib}, our adapter-based models demonstrated superior performance across the 30 LRLs studied, as shown in Table \ref{tab:llms_tc_ner}. Further, our approach outperformed 3-shot results from LLaMA2-7B \cite{llama2}, BLOOM-7B \cite{le2023bloom}, instruction-tuned BLOOMZ-7B \cite{ji2024emma500enhancingmassivelymultilingual}, XGLM \cite{lin2022fewshotlearningmultilinguallanguage}, Occiglot-7B-eu5 \cite{Occiglot}, Yayi \cite{luo2023yayi2multilingualopensource}, LLaMaX2-7B-Alpaca \cite{lu2024llamaxscalinglinguistichorizons}, MaLA-500 \cite{lin2024mala500massivelanguageadaptation}, and recent models like LLaMA3-8B, LLaMA3.1-8B \cite{grattafiori2024llama3herdmodels}, Gemma-7B, Gemma-2-9B \cite{gemmateam2024gemma2improvingopen}, Qwen-1.5-7B, and Qwen-2 \cite{yang2024qwen2technicalreport}. Additionally, our adapter-based approaches surpassed results reported by \citet{asai2023buffet} on the WikiAnn NER task for a subset of 8 overlapping LRLs. Their evaluation included zero- and few-shot prompting with GPT-3.5-Turbo, BLOOM-7B, and instruction-tuned BLOOMZ-7B and mT0-13B \cite{muennighoff2023crosslingualgeneralizationmultitaskfinetuning}. Distilled DeepSeek-R1 models (8B, 14B, 32B, and 70B) \cite{deepseekai2025deepseekr1incentivizingreasoningcapability} 
failed to surpass smaller mLMs on TC.\footnote{Results are zero-shot, with generated token output limited to 100.} Finally, Table \ref{tab:llms_adapter} shows that although \texttt{Seq\_bn\_inv} language-adapter based LLaMA-3-8B improved performance over prompting and its single-task adapter baseline, it was still less effective than smaller mLMs like XLM-R for TC tasks. 

\begin{table}[t]
\centering
\small
\resizebox{\columnwidth}{!}{
\begin{tabular}{@{}lccc@{}}
\toprule
\textbf{Model} & \textbf{TC} ($\uparrow$) & \textbf{SA} ($\uparrow$) & \textbf{NER} ($\uparrow$) \\ 
\midrule
mBERT+\texttt{Seq\_bn\_inv}  & \textbf{71.92} & \textbf{73.68} & \textbf{59.32} \\ 
XLM-R+\texttt{Seq\_bn\_inv}  & \textbf{80.79} & \textbf{83.35} & \textbf{69.26} \\ 
\midrule
LLaMA-3 Baseline  & 31.93 & 58.83 & 45.18 \\ 
LLaMA-3+\texttt{Seq\_bn\_inv}  & 60.26 & 68.68 & 45.12 \\ 
\bottomrule
\end{tabular}
}
\caption{Average F1 scores over 5 selected LRLs for language adapter-tuned LLaMa-3-8B, mBERT, and XLM-R. Additionally, we present results for LLaMA3 with a single \texttt{Seq\_bn} task adapter, similar to our baselines. Per-language results are in Appendix \ref{ap:prompting}.}
\label{tab:llms_adapter}
\end{table}


\section{General Findings and Discussion}
This section highlights key insights gained from our experiments. We analyze performance trends of adapter-based and full fine-tuning approaches for small mLMs, compare their efficacy to LLMs, explore the relationship between language modeling and downstream task performance, and examine the impact of pre- and post-training data sizes on downstream task outcomes.

\subsection{Performance Trends}
\textbf{For MLM, the \texttt{Seq\_bn} adapter consistently achieved the best performance}, likely due to its moderate parameter count (Table \ref{tab:adapter_params} Appendix \ref{app:adapt_params}) aligning with the limited adaptation data. This partially confirms \citet{mundra2024comprehensive}'s findings that simple bottleneck adapters outperform other types, including \texttt{Seq\_bn\_inv} and \texttt{LoRA}. 
Conversely, \texttt{LoRA}, with even fewer parameters, excelled in languages with larger pre-training data in XLM-R, which may reflect that these languages require fewer parameters given their extensive pre-training coverage, considering the limited adaptation data (see Appendix \ref{ap:cor_per}). Moreover, \citet{pfeiffer2020adapterfusion} noted that high-capacity adapters are less effective for XLM-R compared to mBERT. 

\textbf{For downstream tasks, \texttt{Seq\_bn\_inv} slightly outperformed other adapter configurations, with \texttt{Seq\_bn} showing very similar performance in most cases}, confirming findings by \citet{pfeiffer2020mad} that invertible layers enhance adaptation by facilitating input and output embedding alignment. The advantage of \texttt{Seq\_bn\_inv} may also be attributable to its larger number of trainable parameters, which may benefit the task fine-tuning process.
\citet{yong2023bloom1addinglanguagesupport} also report the superiority of using invertible layers for a subset of tested languages on the XNLI task \cite{conneau2018xnli}. Adapter fusion improved NER performance for XLM-R, likely due to the increased count of trainable parameters (compared to individual language adapters), as observed by \citet{lee-etal-2022-fad}. For mBERT, this improvement was not evident: Individual adapters likely provided sufficient capacity.

\textbf{Adapter-based approaches outperformed full fine-tuning for XLM-R and matched mBERT’s performance on MLM, while performing comparably on SA and slightly worse on TC, all with significantly fewer trainable parameters}. This indicates that up to 1 GB of adaptation data suffices for effective adapter training\footnote{This is in line with \citet{bapna2019simplescalableadaptationneural}, \citet{he2021effectivenessadapterbasedtuningpretrained}, and \citet{liu2022fewshotparameterefficientfinetuningbetter}, who report that adapter-based tuning often surpasses full fine-tuning.}, but might be insufficient for fine-tuning larger models like XLM-R.

\textbf{MLM performance} (Tables \ref{tab:mbert_pppl} and \ref{tab:xlm_r_pppl} Appendices \ref{app:mlm_1} and \ref{app:mlm_2}) \textbf{was higher for languages supported by the model's vocabulary.} For unsupported languages in mBERT, such as Sinhala and Amharic
, pseudo-perplexity was artificially low pre-adaptation due to overconfidence in predicting the \texttt{UNK} token. After adaptation, pseudo-perplexity scores increased, reflecting consistent predictions of non-language-specific tokens (e.g., punctuation). Languages with partial script overlap, such as Uyghur and Tibetan, showed minimal improvements. XLM-R's broader script coverage mitigated some issues but still struggled with Tibetan. This highlights the need for vocabulary extension when working with unseen languages \cite{zhang-etal-2020-multi-stage, wang-etal-2020-extending, pfeiffer2021unks}.

\subsection{Small vs. Large LMs for LRLs}
\textbf{Our findings emphasize the effectiveness of adapting smaller mLMs with adapters over relying on prompting or adapting LLMs for LRLs}. The superior performance of smaller mLMs compared to large-scale models has been explored in prior research. \citet{wu2019emerging} observed that limited capacity forces models to align semantically similar representations across languages rather than creating language-specific subspaces. \citet{dufter2020identifying} further showed that overparameterizing mBERT degrades its cross-lingual transfer ability and hypothesized that smaller models produce better language-independent representations by reusing parameters across languages, while larger models tend to partition capacity, limiting shared multilingual representations, later supported by \citet{yong2023bloom1addinglanguagesupport}. Similarly, \citet{shliazhko2023mgptfewshotlearnersmultilingual} found no performance improvements in mGPT when scaling from 1.3B to 13B parameters for classification and factual probing tasks, with mBERT and XLM-R outperforming larger models. Moreover, \citet{pecher2024comparingspecialisedsmallgeneral} noted that larger models do not consistently outperform smaller ones in fine-tuning or prompting settings. These findings, together with our results, collectively argue for prioritizing smaller mLMs over large-scale, resource-intensive models \cite{strubell-etal-2019-energy} to advance performance on LRLs more efficiently and effectively.






\subsection{Correlation Between Language Modeling and Downstream Task Performance}
To investigate the relationship between language modeling and downstream task performance, we performed correlation analyses using Pearson \cite{cohen2009pearson} and Spearman \cite{spearman1961proof} metrics. Results in Table \ref{tab:cor_adapt_ppl} (Appendix \ref{app:cor_ppl_tasks}) show a moderate correlation between pseudo-perplexity and downstream task performance for XLM-R, both pre- and post-adaptation (using Glot data), but a less pronounced correlation for mBERT. \textbf{Lower pseudo-perplexity generally indicated better downstream performance for XLM-R and, to a lesser extent, for mBERT, suggesting its utility as a rough proxy for downstream task capabilities, particularly for larger mLMs}. These findings contrast with prior studies \cite{liang2022holistic, yong2023bloom1addinglanguagesupport}, which reported an unclear relationship between perplexity and task performance.\footnote{Unlike these studies, we evaluate pseudo-perplexity across a diverse set of languages rather than models. This partially aligns with \citet{xia2022training}, who observed a correlation between perplexity and few-shot learning results.} Post-adaptation, the correlation between pseudo-perplexity and downstream performance strengthened, particularly for tasks with consistent data quality (Figure \ref{fig:ppl_base_cor}). We conjecture that the stronger correlations observed for XLM-R likely arise from its optimized multilingual architecture and its extensive pre-training corpus.


\begin{figure*}[t]
    \centering
\includegraphics[width=1.0\textwidth, scale=1.0]{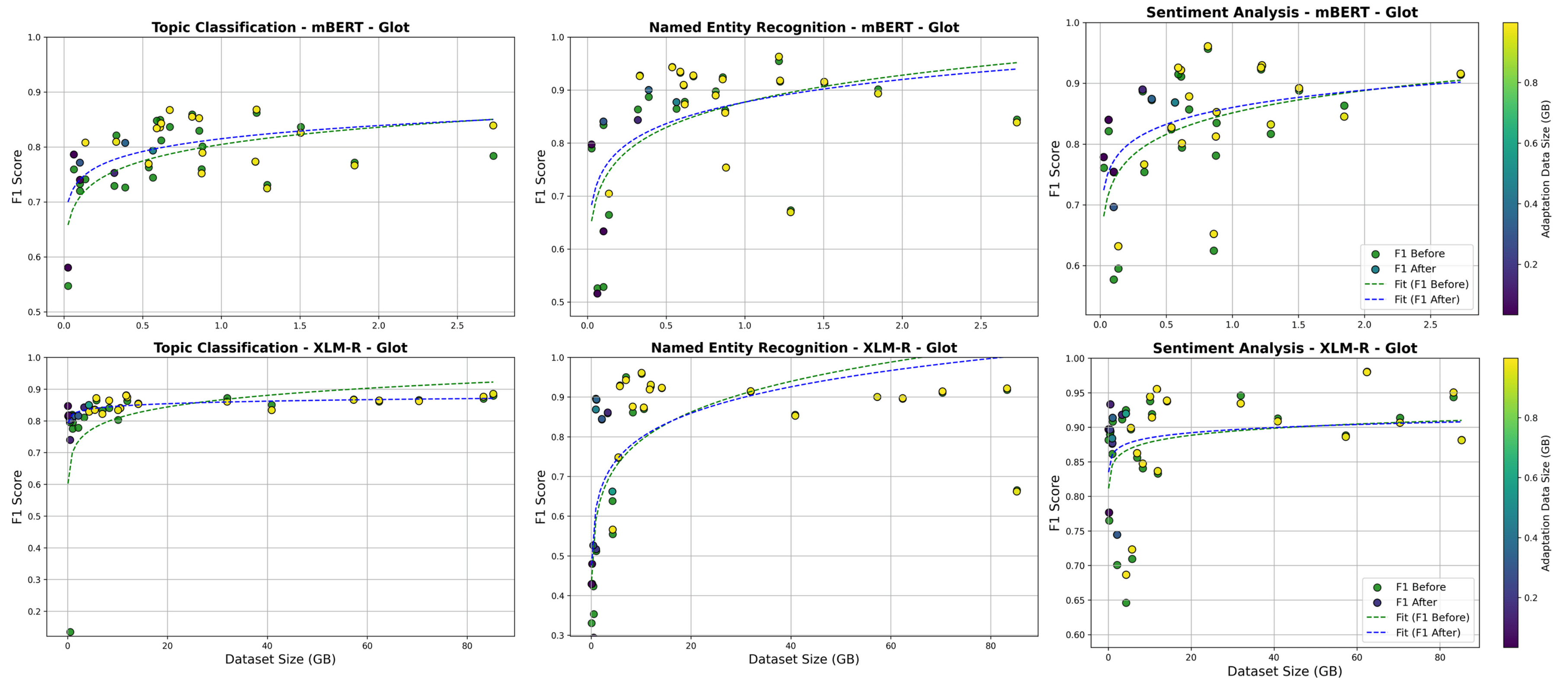}
    \caption{Correlation between the pre-training data sizes for mBERT and XLM-R and downstream task results for the pre-adaptation and post-adaptation results. The vertical bars indicate the amounts of adaptation data. The improvements in downstream performance for both models are primarily concentrated in languages with smaller pre-training data sizes, which are positioned on the left side of the plots. Conversely, for languages with substantial representation in the pre-training data, the improvements are less pronounced or nonexistent (Section \ref{sec:task_data}).}
    \label{fig:data_downstream}
\end{figure*}


\subsection{Impact of Pre- and Post-Training Data Size on MLM and Downstream Tasks} 
\label{sec:task_data}
We analyzed the relationship between pre- and post-adaptation data size and model performance. Before adaptation, pseudo-perplexity and downstream task performance were correlated with pre-training data size (Figure \ref{fig:data_downstream} and Table \ref{tab:cor_ppl_data} Appendix \ref{ap:cor_per}), as also found by \citet{wu-dredze-2020-languages}, \citet{ahuja2023mega} and \citet{bagheri-nezhad-agrawal-2024-drives}. \textbf{Post-adaptation improvements primarily depended on pre-training and, surprisinlgy less so, on adaptation data volumes, with the latter providing only a marginal improvement}.\footnote{Similarly, \citet{kunz2024impactlanguageadapterscrosslingual} show limited overall impact of adaptation data and language adapters.} LRLs exhibited larger gains, while higher-resource languages faced diminishing returns or even reduced performance. The latter is likely due to the model encountering duplicate data already seen during pre-training \cite{lee-etal-2022-deduplicating}. Achieving further gains for well-represented languages may require increasing adaptation data and adapter capacity to better leverage their extensive pre-training coverage. A correlation analysis (Appendix \ref{ap:cor_per}) demonstrates that adaptation data had a stronger impact on mBERT than XLM-R, likely because of its larger relative contribution as compared to pre-training data.

\textbf{In downstream tasks, even small amounts of adaptation data (e.g., a few MB of graph-based data or a few hundred MB of free-text data) produced performance gains}, consistent with \citet{pfeiffer2020mad} and \citet{yong2023bloom1addinglanguagesupport}. This was especially true for mBERT, where adaptation data constitutes a larger proportion relative to its overall training data. For XLM-R, adaptation data was more beneficial for LRLs, while its impact diminished for languages with pre-training data exceeding approximately 20 GB, as also observed by \citet{adelani-etal-2024-sib}. Diminishing returns suggest a threshold effect, where extensive pre-training coverage reduces the utility of adaptation data, indicating that larger adaptation datasets may be necessary for further gains. Figures \ref{fig:tc_vs_data}, \ref{fig:ner_vs_data}, and \ref{fig:sa_vs_data} demonstrate these trends, showing that underrepresented languages typically benefit more from even limited adaptation data, confirmed by correlation analyses (Appendices \ref{app:task_cor_1}, \ref{app:task_cor_2}, and \ref{app:task_cor_3}). 

\textbf{The type of adaptation data influenced task-specific performance.} ConceptNet-based adapters outperformed Glot-based adapters for NER in most languages, likely because ConceptNet contains straightforward NER information. This contrasts with the findings of \citet{gurgurov2024adapting}, who observed different trends when experimenting with a smaller subset of languages. Conversely, Glot-based adapters provided more consistent improvements across tasks, leveraging their larger adaptation data volumes (up to 1 GB for most languages). This emphasizes the important role of relative data size in determining the effectiveness of adaptation across tasks.

\section{Conclusion}
This study evaluated adapter-based adaptation of small mLMs to LRLs using structured and unstructured data, alongside continued pre-training and comparing them with SoTA LLMs. \texttt{Seq\_bn} achieved the best results for MLM tasks, while \texttt{Seq\_bn\_inv} excelled in downstream tasks. Full fine-tuning offered limited advantages over adapters. Downstream performance was primarily influenced by pre-training data, with adaptation data providing incremental gains. Graph-based knowledge from ConceptNet, despite its small size, improved NER performance, while Glot data consistently delivered the largest gains across tasks. Our results generally suggest that smaller mLMs may be better suited for LRLs than LLMs, since mLMs efficiently align cross-lingual representations and generalize well under data constraints. In future research, we intend to investigate how larger models might partition their parameter space across languages and whether they are limiting their ability to leverage shared representations, especially for LRLs.


\section*{Limitations}
This study has three main limitations. First, adapters have specific hyperparameters that influence their behavior and capacity. Future work should systematically explore these hyperparameters and their effects on adapter performance. Second, the amount of adaptation data was limited to 1 GB per language due to computational constraints. Investigating the impact of larger datasets on model adaptation—e.g., utilizing the full GlotCC data without truncation—remains an open and promising direction. Increasing adapter capacity and adaptation data size and measuring adaptation effects as a function of both data volume and model capacity could provide valuable insights. Finally, some experiments were not conducted across all tasks due to resource constraints. For example, adapter fusion was applied only to named entity recognition, and full fine-tuning was only evaluated for small models on masked language modeling, topic classification, and sentiment analysis, but not on named entity recognition.

\section*{Acknowledgments}
This work was supported by DisAI – Improving scientific excellence and creativity in combating disinformation with artificial intelligence and language technologies, a Horizon Europe-funded project under GA No. 101079164, and by the German Ministry of Education and Research (BMBF) as part of the project TRAILS (01IW24005).

\bibliography{custom} 

\appendix \onecolumn

\section*{Appendix}

\section{ConceptNet Tripple Conversion Mapping}
\label{app:cn_conv}

\begin{table}[h!]
\small
\centering
\begin{tabular}{ll}
\toprule
\textbf{ConceptNet Relationship} & \textbf{Natural Language Predicate} \\ 
\midrule
Antonym                         & is the opposite of                   \\ 
DerivedFrom                     & is derived from                      \\ 
EtymologicallyDerivedFrom        & is etymologically derived from       \\ 
EtymologicallyRelatedTo          & is etymologically related to         \\ 
FormOf                           & is a form of                        \\ 
PartOf                           & is a part of                        \\ 
HasA                             & belongs to                          \\ 
UsedFor                          & is used for                         \\ 
AtLocation                       & is a typical location for           \\ 
Causes                           & causes                              \\ 
CausesDesire                     & makes someone want                  \\ 
MadeOf                           & is made of                          \\ 
ReceivesAction                   & receives action of                  \\ 
HasSubevent                      & is a subevent of                    \\ 
HasFirstSubevent                 & is an event that begins with subevent\\ 
HasLastSubevent                  & is an event that concludes with subevent\\ 
HasPrerequisite                  & has prerequisite of                 \\ 
HasProperty                      & can be described as                 \\ 
MotivatedByGoal                  & is a step toward accomplishing the goal\\ 
ObstructedBy                     & is an obstacle in the way of        \\ 
Desires                          & is a conscious entity that typically wants\\ 
CreatedBy                        & is a process or agent that creates  \\ 
CapableOf                        & is capable of                       \\ 
HasContext                       & is a word used in the context of    \\ 
IsA                              & is a type of                        \\ 
RelatedTo                        & is related to                       \\ 
SimilarTo                        & is similar to                       \\ 
Synonym                          & is a synonym of                     \\ 
SymbolOf                         & symbolically represents             \\ 
DefinedAs                        & is a more explanatory version of    \\ 
DistinctFrom                     & is distinct from                    \\ 
MannerOf                         & is a specific way to do             \\ 
LocatedNear                      & is typically found near             \\ 
\bottomrule
\end{tabular}
\caption{ConceptNet relationships and their natural language predicates. This mapping is used for converting the ConceptNet KG data into natural language text.}
\label{tab:relmap}
\end{table}

\newpage
\section{Language Details}
\label{app:lang_det}
\begin{table}[h]

    \centering
    \small
    \def\arraystretch{1.05}
    \resizebox{\columnwidth}{!}{
    \begin{tabular}{l c l l l l c c c c c}
        \toprule
        \textbf{Language} & \textbf{ISO} & \makecell{\textbf{Language} \\ \textbf{Family}} & \textbf{CN (Sent-s)} & \textbf{CN (MB)} & \textbf{Glot (Doc-s)} & \textbf{Glot (MB)} & \textbf{mBERT?} & \textbf{XLM-R?} & \makecell{\textbf{mBERT} \\ \textbf{Data Size (GB)}} & \makecell{\textbf{XLM-R} \\ \textbf{Data Size (GB)}} \\
        \midrule
        Thai & th & Kra-Dai & 123,859 & 6.95 & 2,391,253 & 977.68 & \checkmark & \checkmark & 1.29 & 85.24 \\
        Romanian & ro & Indo-European & 70,236 & 2.47 & 8,657,002 & 1002.36 & \checkmark & \checkmark & 1.22 & 83.29 \\
        Bulgarian & bg & Indo-European & 162,181 & 8.02 & 5,192,702 & 1014.73 & \checkmark & \checkmark & 1.50 & 70.37 \\
        Danish & da & Indo-European & 66,109 & 2.27 & 8,743,767 & 1006.91 & \checkmark & \checkmark & 0.81 & 62.39 \\
        Greek & el & Indo-European & 89,016 & 4.17 & 4,789,519 & 980.94 & \checkmark & \checkmark & 1.85 & 57.30 \\
        Hebrew & he & Afro-Asiatic & 41,444 & 1.62 & 5,287,428 & 991.82 & \checkmark & \checkmark & 2.73 & 40.87 \\
        Slovak & sk & Indo-European & 22,460 & 0.81 & 9,294,165 & 1006.96 & \checkmark & \checkmark & 0.61 & 31.96 \\
        Slovenian & sl & Indo-European & 85,882 & 2.98 & 9,301,902 & 1007.91 & \checkmark & \checkmark & 0.67 & 14.16 \\
        Latvian & lv & Indo-European & 66,408 & 2.4 & 8,301,651 & 988.21 & \checkmark & \checkmark & 0.33 & 11.94 \\
        Indonesian & ms & Austronesian & 175,246 & 6.21 & 8,024,827 & 1022.01 & \checkmark & \checkmark & 0.59 & 11.73 \\
        Georgian & ka & Kartvelian & 35,331 & 1.89 & 3,463,631 & 1014.24 & \checkmark & \checkmark & 0.88 & 10.55 \\
        Bengali & bn & Indo-European & 8,782 & 0.46 & 2,940,197 & 993.44 & \checkmark & \checkmark & 1.22 & 10.10 \\
        Azerbaijani & az & Turkic & 15,149 & 0.57 & 6,179,152 & 1016.68 & \checkmark & \checkmark & 0.62 & 8.33 \\
        Urdu & ur & Indo-European & 13,315 & 0.51 & 4,220,566 & 1009.42 & \checkmark & \checkmark & 0.54 & 6.97 \\
        Macedonian & mk & Indo-European & 38,116 & 1.54 & 5,037,552 & 1005.62 & \checkmark & \checkmark & 0.86 & 5.76 \\
        Telugu & te & Dravidian & 33,476 & 1.72 & 3,162,535 & 1005.55 & \checkmark & \checkmark & 0.88 & 5.46 \\
        Nepali & ne & Indo-European & 4,456 & 0.21 & 2,569,572 & 1012.63 & \checkmark & \checkmark & 0.14 & 4.32 \\
        Marathi & mr & Indo-European & 7,232 & 0.37 & 402,575 & 157.3 & \checkmark & \checkmark & 0.32 & 3.33 \\
        Swahili & sw & Niger-Congo & 12,380 & 0.39 & 2,450,753 & 323.27 & \checkmark & \checkmark & 0.10 & 2.15 \\
        Welsh & cy & Indo-European & 18,313 & 0.61 & 3,174,686 & 360.24 & \checkmark & \checkmark & 0.39 & 1.07 \\
        Uzbek & uz & Turkic & 4,362 & 0.16 & 4,018,172 & 481.49 & \checkmark & \checkmark & 0.57 & 0.95 \\
        Javanese & jv & Austronesian & 3,448 & 0.13 & 367,795 & 43.56 & \checkmark & \checkmark & 0.10 & 0.20 \\
        Sundanese & su & Austronesian & 1,880 & 0.07 & 323,610 & 43.55 & \checkmark & \checkmark & 0.06 & 0.08 \\
        Sinhala & si & Indo-European & 1,782 & 0.1 & 1,655,641 & 586.21 & ✘ & \checkmark & ✘ & 4.27 \\
        Amharic & am & Afro-Asiatic & 1,814 & 0.07 & 667,881 & 203.65 & ✘ & \checkmark & ✘ & 1.00 \\
        Kurdish & ku & Indo-European & 12,246 & 0.44 & 376,260 & 134.7 & ✘ & \checkmark & ✘ & 0.52 \\
        Uyghur & ug & Turkic & 1,715 & 0.06 & 976,010 & 233.61 & ✘ & \checkmark & ✘ & 0.43 \\
        Maltese & mt & Afro-Asiatic & 3,895 & 0.14 & 1,389,527 & 182.17 & ✘ & ✘ & ✘ & ✘ \\
        Tibetan & bo & Sino-Tibetan & 4,768 & 0.21 & 288,847 & 165.31 & ✘ & ✘ & ✘ & ✘ \\
        Yoruba & yo & Niger-Congo & 1,044 & 0.05 & 278,003 & 34.51 & \checkmark & ✘ & 0.03 & ✘ \\
        \bottomrule
    \end{tabular}
    }
    \caption{Number of ConceptNet triples and GlotCC documents as well as corresponding data sizes per language, sorted by Glot (Doc-s) in descending order. The last four columns indicate the inclusion of the respective language in mBERT and XLM-R pre-training data, alongside the corresponding data sizes in GB. The sizes are approximated based on the openly available CC100 and WikiPedia datasets.}
    \label{tab:language-stats}
\end{table}

\newpage
\section{Sentiment Analysis Data Details}
\label{app:sa_data}
\begin{table*}[htb]
\label{ap:sa}
\begin{adjustwidth}{-1cm}{}
    \centering
    \small
    \begin{tabular}{l c | c | c c | c c c}
        \toprule
        \textbf{Language} & \textbf{ISO code} & \textbf{Source} & \textbf{\#pos} & \textbf{\#neg} & \textbf{\#train} & \textbf{\#val} & \textbf{\#test} \\
        \midrule
        Sundanese & su & \citeauthor{winata2022nusax}, \citeyear{winata2022nusax} & 378 & 383 & 381 & 76 & 304 \\
        Amharic & am & \citeauthor{tesfa2024aspect}, \citeyear{tesfa2024aspect} & 487 & 526 & 709 & 152 & 152 \\
        Swahili & sw & \citeauthor{Muhammad2023AfriSentiAT}, \citeyear{Muhammad2023AfriSentiAT}; \citeauthor{muhammad2023semeval}, \citeyear{muhammad2023semeval} & 908 & 319 & 738 & 185 & 304 \\
        Georgian & ka & \citeauthor{stefanovitch-etal-2022-resources}, \citeyear{stefanovitch-etal-2022-resources} & 765 & 765 & 1080 & 120 & 330 \\
        Nepali & ne & \citeauthor{9381292}, \citeyear{9381292} & 680 & 1019 & 1189 & 255 & 255 \\
        Uyghur & ug & \citeauthor{li2022senti}, \citeyear{li2022senti} & 2450 & 353 & 1962 & 311 & 530 \\
        Latvian & lv & \citeauthor{SprogisRikters2020BalticHLT}, \citeyear{SprogisRikters2020BalticHLT} & 1796 & 1380 & 2408 & 268 & 500 \\
        Slovak & sk & \citeauthor{pecar-etal-2019-improving}, \citeyear{pecar-etal-2019-improving} & 4393 & 731 & 3560 & 522 & 1042 \\
        Sinhala & si & \citeauthor{ranathunga2021sentiment}, \citeyear{ranathunga2021sentiment} & 2487 & 2516 & 3502 & 750 & 751 \\
        Slovenian & sl & \citeauthor{buvcar2018annotated}, \citeyear{buvcar2018annotated} & 1665 & 3337 & 3501 & 750 & 751 \\
        Uzbek & uz & \citeauthor{uzbek_sa}, \citeyear{uzbek_sa} & 3042 & 1634 & 3273 & 701 & 702 \\
        Bulgarian & bg & \citeauthor{martinez2021evaluating}, \citeyear{martinez2021evaluating} & 6652 & 1271 & 5412 & 838 & 1673 \\
        Yoruba & yo & \citeauthor{Muhammad2023AfriSentiAT}, \citeyear{Muhammad2023AfriSentiAT}; \citeauthor{muhammad2023semeval}, \citeyear{muhammad2023semeval} & 6344 & 3296 & 5414 & 1327 & 2899 \\
        Urdu & ur & \citeauthor{maas-EtAl:2011:ACL-HLT2011}, \citeyear{maas-EtAl:2011:ACL-HLT2011}; \citeauthor{khan2017harnessing}, \citeyear{khan2017harnessing}; \citeauthor{khan2020usc}, \citeyear{khan2020usc} & 5562 & 5417 & 7356 & 1812 & 1812 \\
        Macedonian & mk & \citeauthor{jovanoski-etal-2015-sentiment}, \citeyear{jovanoski-etal-2015-sentiment} & 3041 & 5184 & 6557 & 729 & 939 \\
        Danish & da & \citeauthor{isbister-etal-2021-stop}, \citeyear{isbister-etal-2021-stop} & 5000 & 5000 & 7000 & 1500 & 1500 \\
        Marathi & mr & \citeauthor{pingle2023l3cube}, \citeyear{pingle2023l3cube} & 5000 & 5000 & 8000 & 1000 & 1000 \\
        Bengali & bn & \citeauthor{sazzed-2020-cross}, \citeyear{sazzed-2020-cross} & 8500 & 3307 & 8264 & 1771 & 1772 \\
        Hebrew & he & \citeauthor{amram-etal-2018-representations}, \citeyear{amram-etal-2018-representations} & 8497 & 3911 & 8932 & 993 & 2483 \\
        Romanian & ro & \citeauthor{tache-etal-2021-clustering}, \citeyear{tache-etal-2021-clustering} & 7500 & 7500 & 10800 & 1200 & 3000 \\
        Telugu & te & \citeauthor{marreddy2022resource}, \citeyear{marreddy2022resource}; \citeauthor{marreddy2022multi}, \citeyear{marreddy2022multi} & 9488 & 6746 & 11386 & 1634 & 3214 \\
        Welsh & cy & \citeauthor{espinosa2021english}, \citeyear{espinosa2021english} & 12500 & 12500 & 17500 & 3750 & 3750 \\
        Azerbaijani & az & \citeauthor{azerbaijanisent}, \citeyear{azerbaijanisent} & 14000 & 14000 & 19600 & 4200 & 4200  \\
        Tibetan & bo & \citeauthor{10348366}, \citeyear{10348366} & 5006 & 5000 & 7004 & 1501 & 1501 \\
        Kurdish & ku & \citeauthor{kurd}, \citeyear{kurd} & 4065 & 3922 & 6000 & 993 & 994  \\
        Greek & el & \citeauthor{kalamatianos2015sentiment}, \citeyear{kalamatianos2015sentiment}; \citeauthor{tsakalidis2018building}, \citeyear{tsakalidis2018building} & 5773 & 1313 & 5936 & 383 & 767  \\
        Javanese & jv & \citeauthor{wongso2021causal}, \citeyear{wongso2021causal} & 12500 & 12500 & 17500 & 5025 & 2475  \\
        Maltese & mt & \citeauthor{dingli2016sentiment}, \citeyear{dingli2016sentiment}; \citeauthor{cortis-davis-2019-social}, \citeyear{cortis-davis-2019-social} & 271 & 580 & 595  & 85 &  171 \\
        Thai & th & \citeauthor{bact_2019_3457447}, \citeyear{bact_2019_3457447}; & 4778 & 6822 & 8103 & 1153 & 2344 \\
        Malay & ms & \citeauthor{purwarianti2019improving}, \citeyear{purwarianti2019improving} & 7319 & 4005 & 7926 & 1132 & 2266  \\
        \bottomrule
    \end{tabular}
    \caption{ Sentiment analysis data details.}
    \label{tab:sa-sources}
\end{adjustwidth}
\end{table*}

\newpage
\section{Named Entity Recognition Data Details}
\label{app:ner_data}
\begin{table*}[htb]
\label{app:ner}
    \centering
    \small
    \begin{tabular}{l c | r r r}
        \toprule
        \textbf{Language} & \textbf{ISO code} & \textbf{\#train} & \textbf{\#val} & \textbf{\#test} \\
        \midrule
        Bulgarian & bg & 20000 & 10000 & 10000 \\
        Indonesian & ms & 20000 & 1000 & 1000 \\
        Maltese & mt & 100 & 100 & 100 \\
        Nepali & ne & 100 & 100 & 100 \\
        Javanese & jv & 100 & 100 & 100 \\
        Uyghur & ug & 100 & 100 & 100 \\
        Tibetan & bo & 100 & 100 & 100 \\
        Sinhala & si & 100 & 100 & 100 \\
        Sundanese & su & 100 & 100 & 100 \\
        Amharic & am & 100 & 100 & 100 \\
        Swahili & sw & 1000 & 1000 & 1000 \\
        Georgian & ka & 10000 & 10000 & 10000 \\
        Latvian & lv & 10000 & 10000 & 10000 \\
        Slovak & sk & 20000 & 10000 & 10000 \\
        Slovenian & sl & 15000 & 10000 & 10000 \\
        Uzbek & uz & 1000 & 1000 & 1000 \\
        Yoruba & yo & 100 & 100 & 100 \\
        Urdu & ur & 20000 & 1000 & 1000 \\
        Macedonian & mk & 10000 & 1000 & 1000 \\
        Danish & da & 20000 & 10000 & 10000 \\
        Marathi & mr & 5000 & 1000 & 1000 \\
        Bengali & bn & 10000 & 1000 & 1000 \\
        Hebrew & he & 20000 & 10000 & 10000 \\
        Romanian & ro & 20000 & 10000 & 10000 \\
        Telugu & te & 1000 & 1000 & 1000 \\
        Welsh & cy & 10000 & 1000 & 1000 \\
        Azerbaijani & az & 10000 & 1000 & 1000 \\
        Greek & el & 20000 & 10000 & 10000 \\
        Kurdish & ku & 100 & 100 & 100 \\
        Thai & th & 20000 & 10000 & 10000 \\
        \bottomrule
    \end{tabular}
    \caption{Named entity recognition data details.}
    \label{tab:ner-sources}
\end{table*}

\newpage
\section{Language Adapters Evaluation Losses}
\label{app:eval_loss}

\begin{table}[h!]
\centering
\small
\def\arraystretch{1.05}
\resizebox{\columnwidth}{!}{
\begin{tabular}{l|cccccc|cccccc}
\toprule
\multirow{2}{*}{\textbf{ISO}} & \multicolumn{6}{c}{\textbf{ConceptNet}} & \multicolumn{6}{c}{\textbf{Glot}} \\
\cmidrule(lr){2-7} \cmidrule(lr){8-13}
 &  \multicolumn{3}{c}{\textbf{mBERT}} & \multicolumn{3}{c}{\textbf{XLM-R}} & \multicolumn{3}{c}{\textbf{mBERT}} & \multicolumn{3}{c}{\textbf{XLM-R}} \\
  \cmidrule(lr){2-4} \cmidrule(lr){5-7} \cmidrule(lr){8-10} \cmidrule(lr){11-13}
 & \texttt{Seq\_bn} & \texttt{LoRA} & \texttt{Seq\_bn\_inv} & \texttt{Seq\_bn} & \texttt{LoRA} & \texttt{Seq\_bn\_inv} & \texttt{Seq\_bn} & \texttt{LoRA} & \texttt{Seq\_bn\_inv} & \texttt{Seq\_bn} & \texttt{LoRA} & \texttt{Seq\_bn\_inv} \\
 \midrule
th & 1.21 & 1.24 & 1.2 & 1.42 & 1.42 & 1.35 & 0.46 & 0.54 & 0.45 & 1.55 & 1.65 & 1.53 \\
ro & 1.41 & 1.46 & 1.34 & 1.43 & 1.43 & 1.33 & 1.37 & 1.52 & 1.34 & 1.27 & 1.3 & 1.26 \\
bg & 0.68 & 0.71 & 0.66 & 0.87 & 0.87 & 0.81 & 1.09 & 1.25 & 1.07 & 1.83 & 1.8 & 1.8 \\
da & 1.24 & 1.29 & 1.19 & 1.35 & 1.36 & 1.26 & 1.39 & 1.54 & 1.36 & 1.28 & 1.36 & 1.26 \\
el & 1.13 & 1.18 & 1.12 & 1.36 & 1.36 & 1.29 & 0.67 & 0.77 & 0.66 & 0.84 & 0.9 & 0.83 \\
he & 1.35 & 1.38 & 1.32 & 1.47 & 1.46 & 1.4 & 1.3 & 1.41 & 1.28 & 1.29 & 1.38 & 1.28 \\
sk & 1.22 & 1.28 & 1.16 & 1.39 & 1.39 & 1.28 & 1.09 & 1.19 & 1.06 & 1.16 & 1.19 & 1.14 \\
sl & 0.83 & 0.91 & 0.79 & 1.05 & 1.09 & 0.98 & 1.16 & 1.28 & 1.13 & 1.22 & 1.28 & 1.21 \\
lv & 1.32 & 1.4 & 1.25 & 1.47 & 1.51 & 1.37 & 1.11 & 1.29 & 1.07 & 1.28 & 1.37 & 1.25 \\
ms & 1.57 & 1.63 & 1.5 & 1.59 & 1.57 & 1.47 & 1.52 & 1.65 & 1.48 & 1.55 & 1.6 & 1.54 \\
ka & 1.15 & 1.19 & 1.14 & 1.38 & 1.35 & 1.3 & 0.79 & 0.91 & 0.77 & 1.12 & 1.18 & 1.11 \\
bn & 0.99 & 1.03 & 0.97 & 1.37 & 1.37 & 1.3 & 1.05 & 1.16 & 1.03 & 1.44 & 1.49 & 1.42 \\
az & 1.33 & 1.37 & 1.29 & 1.5 & 1.55 & 1.42 & 0.89 & 1.02 & 0.86 & 1.19 & 1.31 & 1.15 \\
ur & 1.43 & 1.48 & 1.4 & 1.62 & 1.61 & 1.51 & 1.15 & 1.31 & 1.12 & 1.38 & 1.44 & 1.36 \\
mk & 1.42 & 1.44 & 1.38 & 1.59 & 1.54 & 1.45 & 0.89 & 0.99 & 0.87 & 1.41 & 1.4 & 1.41 \\
te & 1.09 & 1.12 & 1.07 & 1.29 & 1.29 & 1.22 & 0.83 & 0.94 & 0.81 & 1.33 & 1.4 & 1.31 \\
ne & 1.26 & 1.31 & 1.21 & 1.53 & 1.52 & 1.42 & 0.77 & 0.9 & 0.75 & 1.38 & 1.45 & 1.35 \\
mr & 1.08 & 1.12 & 1.04 & 1.46 & 1.45 & 1.37 & 0.94 & 1.07 & 0.92 & 1.43 & 1.49 & 1.41 \\
sw & 1.54 & 1.63 & 1.51 & 1.64 & 1.73 & 1.56 & 0.94 & 1.13 & 0.9 & 1.13 & 1.22 & 1.1 \\
cy & 1.55 & 1.6 & 1.48 & 1.83 & 1.91 & 1.76 & 0.81 & 0.99 & 0.77 & 0.95 & 1.06 & 0.92 \\
uz & 1.22 & 1.3 & 1.18 & 1.55 & 1.62 & 1.45 & 0.85 & 1.01 & 0.82 & 1.06 & 1.17 & 1.03 \\
jv & 1.44 & 1.5 & 1.4 & 1.55 & 1.56 & 1.48 & 2.11 & 2.21 & 2.08 & 2.63 & 2.66 & 2.54 \\
su & 1.51 & 1.56 & 1.47 & 1.38 & 1.4 & 1.38 & 1.14 & 1.28 & 1.11 & 1.21 & 1.35 & 1.18 \\
si & 1.4 & 1.33 & 1.38 & 1.31 & 1.25 & 1.25 & 0.82 & 0.88 & 0.8 & 1.21 & 1.29 & 1.19 \\
am & 1.47 & 1.51 & 1.58 & 1.22 & 1.29 & 1.13 & 1.25 & 1.31 & 1.23 & 1.2 & 1.31 & 1.19 \\
ku & 1.64 & 1.73 & 1.61 & 1.91 & 2.04 & 1.86 & 0.93 & 1.05 & 0.9 & 0.76 & 1.02 & 0.71 \\
ug & 1.09 & 1.13 & 1.07 & 1.57 & 1.59 & 1.47 & 0.46 & 0.57 & 0.44 & 0.79 & 0.94 & 0.76 \\
mt & 1.41 & 1.44 & 1.39 & 1.53 & 1.68 & 1.5 & 0.84 & 1.08 & 0.8 & 0.93 & 1.2 & 0.87 \\
bo & 1.0 & 1.01 & 0.98 & 0.63 & 0.64 & 0.62 & 0.24 & 0.28 & 0.24 & 0.72 & 0.73 & 0.71 \\
yo & 1.12 & 1.27 & 1.1 & 1.77 & 1.79 & 1.76 & 0.87 & 1.04 & 0.84 & 0.83 & 1.03 & 0.78 \\
\bottomrule
\end{tabular}
}
\caption{Evaluation losses for language adapters by model, architecture, and language.}
\label{tab:evaluation_losses}
\end{table}

\textit{Evaluation loss values were not predictive of MLM performance}. Despite \texttt{Seq\_bn\_inv} achieving the lowest evaluation losses, it underperformed in MLM tasks, indicating that evaluation loss may be an unreliable training metric (suggested by \citet{salazar2019masked}).

\newpage
\section{Language Adapter Hyperparameters}
\label{app:adapt_params}





\begin{table}[ht]
\centering
\resizebox{\columnwidth}{!}{
\begin{tabular}{@{}lcccccc|cc}
\toprule
\multirow{2}{*}{\textbf{Adapter Type}} & \multicolumn{3}{c}{\textbf{mBERT}}                       & \multicolumn{3}{c}{\textbf{XLM-R}}                       & \multicolumn{2}{c}{\textbf{LLaMA-3}} \\ 
\cmidrule(lr){2-4} \cmidrule(lr){5-7} \cmidrule(lr){8-9} & \textbf{\texttt{Seq\_bn}} & \textbf{\texttt{Seq\_bn\_inv}} & \textbf{\texttt{LoRA}} & \textbf{\texttt{Seq\_bn}} & \textbf{\texttt{Seq\_bn\_inv}} & \textbf{\texttt{LoRA}} & \textbf{\texttt{Seq\_bn}} & \textbf{\texttt{Seq\_bn\_inv}} \\ 
\midrule
Trainable Params (No.)                 & 894,528          & 1,190,592             & 294,912       & 894,528          & 1,190,592             & 294,912   &  67,248,128     & 75,642,880             \\
Trainable Params (\%)                  & 0.505\%          & 0.672\%               & 0.166\%       & 0.322\%          & 0.429\%               & 0.106\%    & 0.896\%   & 1.008\%               \\
\midrule

\textbf{Hyperparameters for LA}               & \multicolumn{6}{c}{\begin{tabular}[c]{@{}c@{}}Batch Size: 16, Learning Rate: 1e-4, \\ \texttt{Seq\_bn} and \texttt{Seq\_bn\_inv}: Reduction Factor = 16, \\ \texttt{LoRA}: $\alpha = 8$, $r = 8$\end{tabular}} & \multicolumn{2}{c}{\begin{tabular}[c]{@{}c@{}}Batch Size: 1, \\ Learning Rate: 1e-4\end{tabular}} \\ 
\textbf{Hyperparameters for TA}               & \multicolumn{6}{c}{\begin{tabular}[c]{@{}c@{}}Batch Size: 32, Learning Rate: 1e-4, \\ \texttt{Seq\_bn}: Reduction Factor = 16, \\ \texttt{LoRA}: $\alpha = 8$, $r = 8$\end{tabular}} & \multicolumn{2}{c}{\begin{tabular}[c]{@{}c@{}}Batch Size for TC: 16; for SA and NER: 8, \\ Learning Rate: 2e-5\end{tabular}} \\ 
\bottomrule
\end{tabular}
}
\caption{Trainable parameters and hyperparameters for different adapter types in mBERT, XLM-R, and LLaMA-3. The rest of hyperparameters are as specified in the default adapter configurations in Adapterhub. LA - Langauge adapter, TA - Task adapter.}

\label{tab:adapter_params}
\end{table}

\newpage
\section{Masked Language Modeling Pseudo-Perplexity - Part I}
\label{app:mlm_1}
\begin{table}[h]
\centering
\small
\begin{tabular}{l|c:ccc|ccc|c}
\toprule
\multirow{2}{*}{\textbf{ISO}} & \multicolumn{8}{c}{\textbf{mBERT}} \\
\cmidrule(lr){2-9}
 & \multicolumn{1}{c}{} & \multicolumn{3}{c}{\textbf{ConceptNet}} & \multicolumn{4}{c}{\textbf{Glot}} \\
 \cmidrule(lr){3-5} \cmidrule(lr){6-9}
 & \textbf{Base} & \textbf{\texttt{Seq\_bn}} & \textbf{\texttt{LoRA}} & \textbf{\texttt{Seq\_bn\_inv}} & \textbf{\texttt{Seq\_bn}} & \textbf{\texttt{LoRA}} & \textbf{\texttt{Seq\_bn\_inv}} & \textbf{FFT} \\
\midrule
he & 18.36 & 19.71 & 18.29 & 19.85 & \textbf{11.09} & 12.31 & 12.51 & \underline{8.78} \\
el & 4.69 & 6.17 & 5.55 & 6.92 & \textbf{3.3} & 3.54 & 3.49 & \underline{2.71} \\
bg & 10.84 & 14.99 & 12.65 & 20.93 & \textbf{5.4} & 5.9 & 6.09 & \underline{4.67} \\
th & 3.87 & 4.13 & 4.29 & 4.07 & \textbf{2.94} & 3.34 & 3.18 & \underline{2.54} \\
ro & 11.49 & 13.47 & 12.67 & 22.39 & \textbf{5.94} & 6.59 & 8.67 & 6.75 \\
bn & 11.97 & 14.94 & 13.53 & 15.99 & \textbf{9.11} & 10.05 & 10.32 & \underline{8.42} \\
te & 7.92 & 8.9 & 8.34 & 9.33 & \textbf{6.09} & 6.13 & 6.4 & \underline{5.32} \\
ka & 6.52 & 6.3 & 6.0 & 6.54 & \textbf{3.63} & 4.06 & 3.91 & \underline{2.6} \\
mk & 11.95 & 14.5 & 12.3 & 13.26 & \textbf{5.83} & 6.33 & 6.54 & \underline{5.53} \\
da & 19.16 & 19.29 & 25.39 & 30.87 & \textbf{11.13} & 11.8 & 13.02 & \underline{8.76} \\
sl & 13.57 & 18.09 & 14.32 & 26.86 & \textbf{6.68} & 7.26 & 8.58 & \underline{4.91} \\
az & 12.47 & 15.2 & 13.48 & 24.26 & \textbf{7.04} & 7.89 & 7.9 & \underline{5.83} \\
sk & 11.5 & 13.86 & 12.37 & 19.29 & \textbf{5.98} & 6.64 & 7.14 & 6.03 \\
ms & 36.26 & 53.66 & 50.17 & 128.6 & \textbf{18.23} & 20.01 & 22.71 & \underline{16.95} \\
uz & 26.65 & 31.41 & 23.43 & 40.35 & \textbf{5.84} & 7.21 & 9.22 & \underline{3.84} \\
ur & 22.59 & 23.02 & 21.74 & 26.4 & \textbf{10.18} & 12.0 & 12.89 & \underline{7.16} \\
cy & 21.24 & 22.13 & 23.0 & 39.75 & \textbf{6.08} & 7.8 & 9.06 & \underline{4.89} \\
lv & 14.14 & 18.31 & 16.21 & 33.14 & \textbf{5.98} & 7.13 & 7.48 & \underline{4.58} \\
mr & 12.51 & 12.9 & 12.21 & 14.0 & \textbf{5.84} & 6.78 & 6.85 & 6.71 \\
ne & 12.72 & 14.19 & 13.08 & 15.36 & \textbf{6.71} & 7.21 & 8.68 & \underline{4.88} \\
jv & 83.84 & 115.27 & 132.08 & 146.64 & \textbf{19.4} & 22.86 & 31.6 & \underline{19.19} \\
sw & 42.53 & 57.57 & 52.21 & 79.5 & \textbf{8.99} & 12.48 & 16.09 & \underline{7.19} \\
su & 102.16 & 177.27 & 183.04 & 227.87 & \textbf{20.24} & 23.2 & 34.29 & 34.93 \\
yo & 85.21 & 293.99 & 210.43 & 370.71 & \textbf{23.14} & 31.96 & 86.79 & 38.89 \\
\hdashline
Avg. & 25.17 & 41.22 & 37.37 & 55.95 & \textbf{8.95} & 10.44 & 14.31 & 9.25 \\
\midrule
mt$^{\dagger}$ & 531.59 & 432.99 & 456.64 & 457.43 & \textbf{6.89} & 9.87 & 15.02 & \underline{5.95} \\
ku$^{\dagger}$ & \textbf{72.87} & 119.29 & 101.13 & 149.74 & 1524.98 & 559.83 & 173.24 & 6381.75 \\
ug$^{\dagger}$ & 112.63 & 96.52 & 86.31 & 121.15 & \textbf{28.69} & 67.26 & 75.53 & 313.64 \\
si$^{\dagger}$ & \textbf{16.29} & 96.5 & 40.3 & 103.36 & 15640.68 & 8981.09 & 157397.73 & 443921.11 \\
am$^{\dagger}$ & \textbf{10.06} & 31.41 & 26.93 & 23.47 & 56052.75 & 34924.59 & 4223.4 & 38289.93 \\
bo$^{\dagger}$ & \textbf{4.59} & 58.78 & 47.33 & 89.81 & 57.94 & 65.03 & 1136.47 & 41.99 \\
\hdashline
Avg. & \textbf{124.67} & 139.25 & 126.44 & 157.49 & 12218.65 & 7434.61 & 27170.23 & 81492.4 \\
\midrule
Total & \textbf{45.07} & 60.83 & 55.18 & 76.26 & 2450.89 & 1495.27 & 5445.49 & 16305.88 \\
\bottomrule
\end{tabular}
\caption{Pseudo-perplexity scores comparison across different adapters for mBERT in ConceptNet and Glot. $^{\dagger}$Language not included in mBERT pre-training. FFT denotes full fine-tuning of a base model on the target-language Glot data. The underlined FFT scores indicate that FFT outperform the best performing adapter for a respective language.}
\label{tab:mbert_pppl}
\end{table}

\newpage
\section{Masked Language Modeling Pseudo-Perplexity - Part II}
\label{app:mlm_2}
\begin{table}[h]
\centering
\small
\def\arraystretch{1.05}
\begin{tabular}{l|c:ccc|ccc|c}
\toprule
\multirow{2}{*}{\textbf{ISO}} & \multicolumn{8}{c}{\textbf{XLM-R}} \\
\cmidrule(lr){2-9}
 &  \multicolumn{1}{c}{} & \multicolumn{3}{c}{\textbf{ConceptNet}} & \multicolumn{4}{c}{\textbf{Glot}} \\
\cmidrule(lr){3-5} \cmidrule(lr){6-9}
& \textbf{Base} & \textbf{\texttt{Seq\_bn}} & \textbf{\texttt{LoRA}} & \textbf{\texttt{Seq\_bn\_inv}} & \textbf{\texttt{Seq\_bn}} & \textbf{\texttt{LoRA}} & \textbf{\texttt{Seq\_bn\_inv}} & \textbf{FFT} \\
 \midrule
th & \textbf{7.83} & 8.67 & 8.86 & 10.11 & 8.78 & 7.97 & 9.39 & 22.16 \\
ro & 2.97 & 3.76 & 3.79 & 4.51 & 3.42 & \textbf{2.96} & 3.25 & 6.18 \\
bg & \textbf{3.61} & 4.88 & 5.51 & 5.4 & 3.63 & 3.7 & 3.64 & 6.12 \\
da & 4.29 & 5.56 & 5.94 & 6.21 & 6.69 & \textbf{4.21} & 4.58 & 7.9 \\
el & \textbf{2.56} & 3.17 & 3.1 & 3.46 & 2.97 & 2.63 & 2.87 & 3.81 \\
he & \textbf{5.74} & 6.17 & 6.36 & 6.74 & 5.8 & 5.84 & 5.99 & 10.95 \\
sk & 3.93 & 4.85 & 4.67 & 5.36 & 4.56 & \textbf{3.68} & 4.08 & 4.62 \\
sl & 4.79 & 7.31 & 7.41 & 8.68 & 4.35 & \textbf{4.01} & 4.95 & 5.3 \\
lv & 4.14 & 5.96 & 6.32 & 9.34 & 5.09 & \textbf{3.92} & 4.7 & \underline{4.87} \\
ms & 10.79 & 15.02 & 15.82 & 17.26 & 8.97 & \textbf{8.8} & 9.65 & 12.55 \\
ka & \textbf{3.88} & 4.41 & 4.47 & 4.48 & 3.99 & 3.94 & 4.76 & 4.97 \\
bn & 6.5 & 7.22 & 7.17 & 7.6 & \textbf{5.95} & 6.28 & 8.0 & 6.69 \\
az & 7.52 & 11.21 & 11.45 & 15.95 & 8.27 & \textbf{7.58} & 9.7 & 14.11 \\
ur & 10.17 & 12.13 & 12.82 & 12.23 & \textbf{9.53} & 9.54 & 11.12 & 12.32 \\
mk & 5.19 & 6.74 & 7.51 & 7.28 & 4.82 & \textbf{4.78} & \textbf{4.78} & 8.14 \\
te & 6.76 & 8.12 & 8.11 & 8.31 & \textbf{6.41} & 6.66 & 9.92 & 7.6 \\
ne & 12.76 & 16.87 & 17.74 & 16.91 & 11.86 & \textbf{11.82} & 22.42 & 16.64 \\
si & 7.04 & 7.97 & 8.22 & 8.26 & \textbf{5.74} & 6.37 & 11.44 & 6.74 \\
mr & 10.25 & 11.83 & 12.12 & 12.67 & 9.11 & \textbf{8.9} & 16.42 & 21.99 \\
sw & 15.68 & 26.99 & 27.39 & 36.78 & \textbf{7.76} & 9.61 & 11.24 & 9.18 \\
cy & 9.37 & 13.94 & 16.05 & 17.51 & \textbf{5.08} & 5.88 & 8.11 & \underline{4.7} \\
am & 10.87 & 14.77 & 15.4 & 15.15 & \textbf{7.32} & 8.44 & 17.0 & 10.49 \\
uz & 8.4 & 14.77 & 16.81 & 20.66 & \textbf{5.46} & 6.21 & 9.14 & 5.92 \\
ku & 159.39 & 72.75 & 84.04 & 69.25 & \textbf{2.95} & 4.34 & 19.27 & 3.88 \\
ug$^{\dagger}$ & 6.87 & 13.97 & 12.48 & 16.76 & \textbf{4.99} & 5.97 & 12.48 & 16.13 \\
jv & 33.81 & 96.45 & 89.36 & 116.95 & \textbf{12.49} & 15.06 & 27.14 & 26.25 \\
su & 57.32 & 134.71 & 128.95 & 152.14 & \textbf{10.41} & 15.22 & 29.1 & 25.16 \\
\hdashline
Avg. & 15.65 & 20.01 & 20.29 & 22.81 & \textbf{6.53} & 6.83 & 10.56 & 10.57 \\
\midrule
mt$^{\ddagger}$ & 395.18 & 283.77 & 335.23 & 275.56 & \textbf{3.19} & 5.0 & 12.01 & 3.36 \\
bo$^{\ddagger}$ & \textbf{9.45} & 937.1 & 2036.45 & 1209.39 & 353.49 & 274.66 & 1972.96 & 597.55 \\
yo$^{\ddagger}$ & 207.26 & 225.8 & 335.24 & 223.49 & \textbf{9.57} & 14.31 & 155.99 & 19.12 \\
\hdashline
Avg. & 203.96 & 482.22 & 902.31 & 569.48 & 122.08 & \textbf{97.99} & 713.65 & 206.68 \\
\midrule
Total & 34.48 & 66.23 & 108.49 & 77.48 & 18.09 & \textbf{15.94} & 80.87 & 30.18 \\
\bottomrule
\end{tabular}
\caption{Pseudo-perplexity scores comparison for XLM-R across different adapters in ConceptNet and Glot. $^{\ddagger}$Language not included in XLM-R pre-training. FFT denotes full fine-tuning of a base model on the target-language Glot data. The underlined FFT scores indicate that FFT outperform the best performing adapter for a respective language.}
\label{tab:xlm_r_pppl}
\end{table}

\newpage
\section{Correlation Between Pseudo-Perplexity Pre- and Post-training Data Sizes}
\label{ap:cor_per}

\begin{figure}[h]
    \centering
    \includegraphics[width=1.0\textwidth]{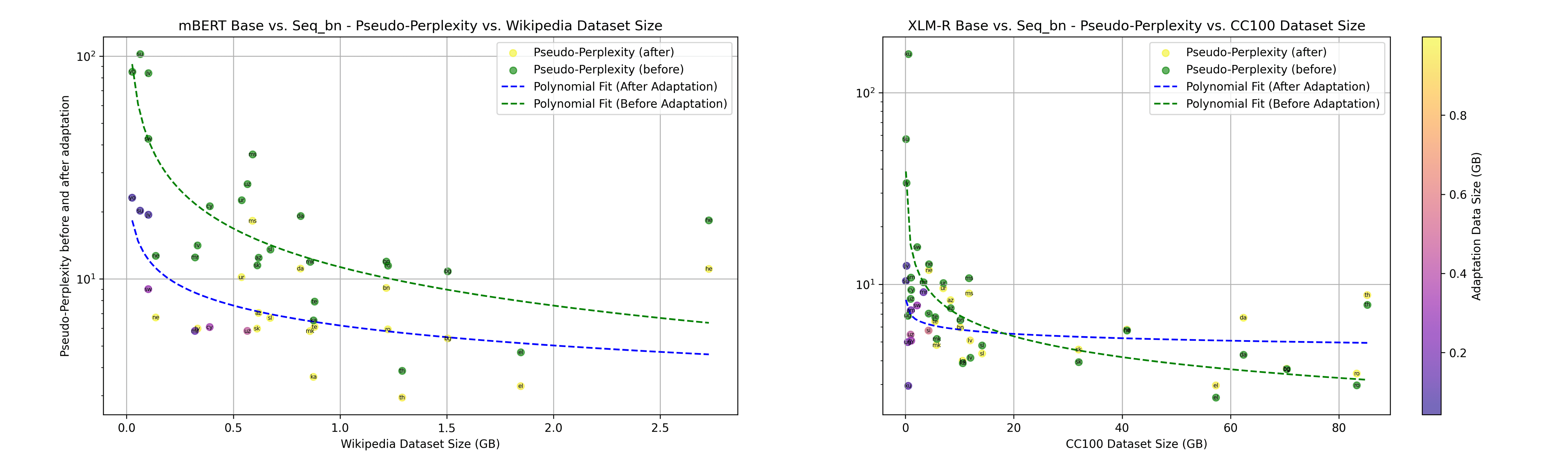}
    \caption{Correlation between the pre-training data sizes for mBERT and XLM-R and the pseudo-perplexities with the values fit in the log-space for the pre-adaptation and post-adaptation results.}
    \label{fig:ppl_size}
\end{figure}

\begin{table}[ht]
\centering
\small
\begin{tabular}{cccccc}
\toprule
\textbf{} & \textbf{Model} & \textbf{Pearson (p-value)} & \textbf{Spearman (p-value)} \\
\midrule
\multirow{2}{*}{\rotatebox{25}{\textbf{\textit{Pre-adapt}}}} 
& mBERT & -0.37 (\textit{0.07}) & -0.51 (\textit{0.01}) \\
& XLM-R & -0.32 (\textit{0.1}) & -0.39 (\textit{0.04}) \\
\midrule\midrule
\multirow{2}{*}{\rotatebox{25}{\textbf{\textit{Post-adapt}}}} 
& mBERT & -0.69 (\textit{<0.001}) & -0.79 (\textit{<0.001}) \\
& XLM-R & -0.27 (\textit{0.16}) & -0.79 (\textit{<0.001}) \\
\bottomrule
\end{tabular}
\caption{Pearson and Spearman Correlations for mBERT and XLM-R between pseudo-perplexity and amounts of pre-training and post-training data for the pre-adaptation and post-adaptation results. Post-adaptation results are based on the models with Seq\_bn language adapters and denote the correlation between the sum of the pre-training and adaptation data sizes and pseudo-perplexity scores after the adaptation.}
\label{tab:cor_ppl_data}
\end{table}

As illustrated in Figure \ref{fig:ppl_size}, the improvements in pseudo-perplexity for both models are primarily concentrated in languages with smaller pre-training data sizes, which are positioned on the left side of the plots. These languages benefit the most from the adaptation process. Conversely, for languages with substantial representation in the pre-training data, the improvements are less pronounced or nonexistent. \textit{This suggests that underrepresented languages in the pre-training data can achieve significant gains in pseudo-perplexity even with modest amounts of adaptation data and low-capacity adapters (smaller parameter counts). In contrast, further improvements for well-represented languages may require increasing the capacity of the adapters to better utilize their substantial pre-training representation.} The stagnation, or drops, in the performance on the languages with extensive pre-training data effects can also be attributed to the model seeing the same (duplicated) data that was seen during pre-training, which makes the "value" of data lower since the model sees the duplicates \cite{lee-etal-2022-deduplicating}.

\newpage
\section{Comparison of XLM-R-base with Glot500 and XLM-R-large}
\label{ap:adaptervsbigmodels}
\begin{table}[h!]
\centering
\small
\begin{tabular}{l|c|c|c|c}
\toprule
\textbf{ISO} & \textbf{XLM-R-base} & \textbf{Adapted XLM-R-base} & \textbf{XLM-R-large} & \textbf{Glot-500m} \\
\midrule
th & 7.83 & 7.97 & \textbf{4.92} & 31.34 \\
ro & 2.97 & 2.96 & \textbf{2.06} & 13.29 \\
bg & 3.61 & 3.63 & \textbf{2.53} & 14.16 \\
da & 4.29 & 4.21 & \textbf{2.78} & 28.06 \\
el & 2.56 & 2.97 & \textbf{1.87} & 6.87 \\
he & 5.74 & 5.8 & \textbf{3.19} & 32.80 \\
sk & 3.93 & 3.68 & \textbf{2.30} & 26.36 \\
sl & 4.79 & 4.01 & \textbf{2.60} & 41.98 \\
lv & 4.14 & 3.92 & \textbf{2.51} & 14.55 \\
ms & 10.79 & 8.8 & \textbf{6.71} & 38.46 \\
ka & 3.88 & 3.94 & \textbf{2.69} & 10.77 \\
bn & 6.50 & 5.95 & \textbf{3.99} & 19.36 \\
az & 7.52 & 7.58 & \textbf{4.40} & 17.46 \\
ur & 10.17 & 9.53 & \textbf{6.10} & 25.60 \\
mk & 5.19 & 4.78 & \textbf{3.23} & 14.00 \\
te & 6.76 & 6.41 & \textbf{4.31} & 17.19 \\
ne & 12.76 & 11.82 & \textbf{8.06} & 23.19 \\
mr & 10.25 & 8.9 & \textbf{5.77} & 27.95 \\
sw & 15.68 & \textbf{7.76} & 8.90 & 44.82 \\
cy & 9.37 & 5.08 & \textbf{4.35} & 25.74 \\
uz & 8.40 & 5.46 & \textbf{3.92} & 15.33 \\
jv & 33.81 & \textbf{12.49} & 17.83 & 73.46 \\
su & 57.32 & \textbf{10.41} & 26.42 & 52.65 \\
si & 7.04 & 5.74 & \textbf{4.50} & 15.03 \\
am & 10.87 & 7.32 & \textbf{6.73} & 25.56 \\
ku & 159.39 & \textbf{2.95} & 126.40 & 23.35 \\
ug & 6.87 & 4.99 & \textbf{3.80} & 13.67 \\
\hdashline
Avg. & 15.65 & \textbf{6.26} & 10.11 & 25.66 \\
\midrule
mt & 395.18 & \textbf{3.19} & 317.81 & 7.93 \\
bo & 9.45 & 274.66 & \textbf{3.99} & 26.74 \\
yo & 207.26 & \textbf{9.57} & 155.57 & 96.80 \\
\hdashline
Avg. & 203.96 & 95.81 & 159.12 & \textbf{43.82} \\
\midrule
Total & 34.48 & \textbf{15.22} & 25.01 & 27.48 \\
\bottomrule
\end{tabular}
\caption{Average pseudo-perplexity scores for 30 languages across three model configurations. For the adapted XLM-R-base, we pick the adapter with the best performance.}
\label{tab:adaptervsbigmodels}
\end{table}

We additionally compare XLM-R adapted with Glot language adapters against two larger models: XLM-R-large \cite{conneau2019unsupervised} and Glot500-m \cite{imanigooghari-etal-2023-glot500} (Table \ref{tab:adaptervsbigmodels}). Both models provide distinct points of comparison. XLM-R-large shares the same architecture as XLM-R-base but with a significantly larger size (550M parameters). XLM-R-large outperformed smaller models with adapters on MLM, suggesting that adapter effectiveness might be inherently constrained by the base model’s capacity.  In contrast, Glot500-m, while only slightly larger than XLM-R-base (395M parameters), introduces an extended vocabulary to support new scripts from a 600GB multilingual corpus and fine-tunes the weights of XLM-R-base. Its training employs a sampling strategy with an alpha of 0.3, prioritizing low-resource languages over high-resource ones. While this approach improves its performance on many low-resource languages, it results in suboptimal outcomes for well-represented languages.

This comparison is particularly relevant as it evaluates whether fine-tuning XLM-R-base with Glot-based language adapters can surpass the performance of these larger models. Furthermore, Glot500-m provides a unique benchmark, as it was trained on the same multilingual corpus used for our adapters, albeit without the computational constraints that limited our data size for adaptation.

\newpage
\section{Correlation Between Pseudo-Perplexity and Downstream Tasks}
\label{app:cor_ppl_tasks}

\begin{figure}[hbtp] 
    \centering
    \includegraphics[width=1.0\textwidth]{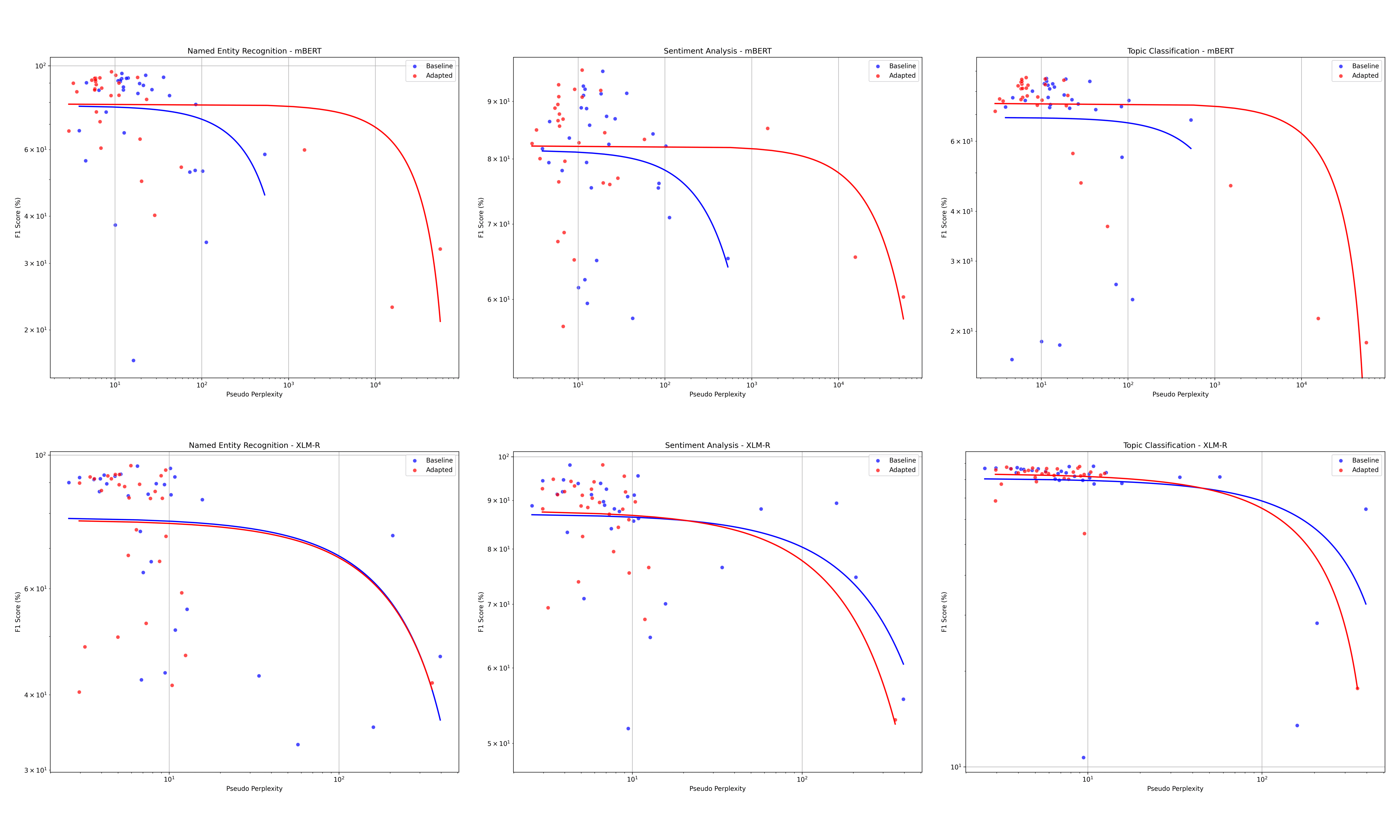}
    \caption{Correlation between the downstream performance for mBERT and XLM-R pre- and post-adaptation and the pseudo-perplexities.}
    \label{fig:ppl_base_cor}
\end{figure}

\begin{table}[h]
\centering
\small
\begin{tabular}{cc|cc:cccc}
\toprule
\textbf{Model} & \textbf{Task} & \multicolumn{2}{c|}{\textbf{Pre-Adapt}} & \multicolumn{2}{c}{\textbf{Post-Adapt}} \\
\cmidrule{3-6}
& & \textbf{Pearson (p-value)} & \textbf{Spearman (p-value)} & \textbf{Pearson (p-value)} & \textbf{Spearman (p-value)} \\
\midrule
\multirow{3}{*}{mBERT} 
& TC & -0.09 (\textit{0.62}) & -0.25 (\textit{0.18}) & -0.66 (\textit{<0.001}) & -0.42 (\textit{0.02}) \\
& SA & -0.29 (\textit{0.12}) & -0.15 (\textit{0.42}) & -0.45 (\textit{0.01}) & -0.23 (\textit{0.23}) \\
& NER & -0.28 (\textit{0.13}) & -0.22 (\textit{0.24}) & -0.54 (\textit{0.002}) & -0.49 (\textit{0.006}) \\
\midrule
\multirow{3}{*}{XLM-R} 
& TC & -0.48 (\textit{0.007}) & -0.68 (\textit{<0.001}) & -0.88 (\textit{<0.001}) & -0.20 (\textit{0.3}) \\
& SA & -0.47 (\textit{0.009}) & -0.55 (\textit{0.002}) & -0.64 (\textit{<0.001}) & -0.38 (\textit{0.04}) \\
& NER & -0.42 (\textit{0.02}) & -0.62 (\textit{<0.001}) & -0.35 (\textit{0.06}) & -0.28 (\textit{0.13}) \\
\bottomrule
\end{tabular}
\caption{Pearson and Spearman Correlations for mBERT and XLM-R (Pre-Adapt and Post-Adapt) between pseudo-perplexity and task performance. Post-Adapt is represented by the models adapted with the Seq\_bn language adapters.}
\label{tab:cor_adapt_ppl}
\end{table}

\newpage
\section{Topic Classification Results - Part I}
\label{app:tc_1}

\begin{table}[h!]
\centering
\small
\def\arraystretch{1.05}
\begin{tabular}{l|c:ccc|ccc|c}
\toprule
\multirow{2}{*}{\textbf{ISO}} & \multicolumn{8}{c}{\textbf{mBERT}} \\
\cmidrule(lr){2-9}
 &  \multicolumn{1}{c}{} & \multicolumn{3}{c}{\textbf{ConceptNet}} & \multicolumn{4}{c}{\textbf{Glot}} \\
\cmidrule(lr){3-5} \cmidrule(lr){6-9}
& \textbf{Base} & \textbf{\texttt{Seq\_bn}} & \textbf{\texttt{LoRA}} & \textbf{\texttt{Seq\_bn\_inv}} & \textbf{\texttt{Seq\_bn}} & \textbf{\texttt{LoRA}} & \textbf{\texttt{Seq\_bn\_inv}} & \textbf{FFT} \\
\midrule
he & 79.79 & \textbf{83.99} & 82.87 & 82.11 & 83.26 & 83.43 & 83.91 & 83.24 \\
el & \textbf{79.47} & 77.95 & 79.14 & 78.12 & 76.65 & 77.92 & 76.64 & \underline{84.81} \\
bg & \textbf{84.39} & 83.71 & 84.17 & 83.38 & 82.64 & 82.87 & 82.58 & \underline{85.88} \\
th & 74.18 & \textbf{74.66} & 73.9 & 74.42 & 71.34 & 74.47 & 72.47 & \underline{76.44} \\
ro & 86.95 & 87.86 & 86.45 & \textbf{88.37} & 85.8 & 86.63 & 86.8 & \underline{89.06} \\
bn & 76.18 & 77.65 & 74.52 & 76.69 & 77.51 & \textbf{78.09} & 77.34 & 77.3 \\
te & 80.03 & \textbf{82.35} & 80.04 & 81.13 & 77.32 & 81.2 & 78.95 & 79.33 \\
ka & 76.28 & 73.26 & 74.26 & 74.07 & 75.68 & \textbf{78.23} & 75.19 & \underline{79.82} \\
mk & 83.44 & 84.48 & 84.34 & 83.79 & 84.53 & 84.92 & \textbf{85.25} & 84.96 \\
da & 87.06 & 86.85 & 86.63 & \textbf{87.72} & 86.03 & 86.48 & 85.5 & 85.8 \\
sl & 83.6 & 85.07 & 83.75 & 86.22 & 86.71 & 85.39 & \textbf{86.73} & 86.43 \\
az & 81.09 & 83.72 & 82.53 & 83.38 & 82.93 & 82.55 & \textbf{84.29} & 82.01 \\
sk & 84.37 & 83.49 & 83.98 & \textbf{85.4} & 84.79 & 84.43 & 83.57 & 84.52 \\
ms & 84.31 & 84.65 & 84.1 & 82.94 & \textbf{85.4} & 84.59 & 83.39 & 84.38 \\
uz & 76.57 & 73.89 & 73.71 & 75.76 & \textbf{81.32} & 74.44 & 79.35 & \underline{85.35} \\
ur & 76.7 & 73.7 & 74.85 & 74.76 & 76.06 & 75.26 & \textbf{76.94} & \underline{78.18} \\
cy & 72.37 & 72.23 & 71.6 & 73.49 & \textbf{81.47} & 77.16 & 80.75 & \underline{85.53} \\
lv & 82.28 & \textbf{83.63} & 82.42 & 82.45 & 83.48 & 82.56 & 80.94 & \underline{85.02} \\
mr & 73.21 & \textbf{77.29} & 76.22 & 76.61 & 76.37 & 75.73 & 75.28 & \underline{78.84} \\
ne & 73.72 & 77.55 & 74.62 & 76.02 & \textbf{81.59} & 75.21 & 80.8 & 79.11 \\
jv & 72.4 & 73.32 & \textbf{75.12} & 73.11 & 73.71 & 74.09 & 74.02 & \underline{75.89} \\
sw & 69.17 & 70.53 & 69.89 & 70.21 & 73.93 & 69.05 & \textbf{77.15} & \underline{85.89} \\
su & 76.15 & 77.42 & 77.62 & 77.0 & 78.21 & \textbf{79.2} & 78.63 & \underline{79.97} \\
yo & 54.18 & 52.11 & 52.08 & 54.89 & 55.93 & 55.93 & \textbf{58.05} & \underline{63.66} \\
\hdashline
Avg. & 77.67 & 78.39 & 77.87 & 78.42 & 79.28 & 78.74 & \textbf{79.35} & \underline{81.73} \\
\midrule
mt$^{\dagger}$ & 69.86 & 69.83 & 69.85 & 68.79 & 78.0 & 78.09 & \textbf{79.8} & \underline{83.32} \\
ku$^{\dagger}$ & 28.76 & 23.78 & 15.71 & 19.93 & 46.41 & 40.22 & \textbf{46.85} & \underline{52.82} \\
ug$^{\dagger}$ & 23.4 & 22.21 & 20.9 & 22.17 & 47.18 & 31.68 & \textbf{48.91} & \underline{56.26} \\
si$^{\dagger}$ & 17.45 & 14.3 & 14.88 & 14.95 & \textbf{21.53} & 21.25 & 20.4 & 19.08 \\
am$^{\dagger}$ & 17.75 & 14.01 & 18.47 & 12.94 & 18.74 & \textbf{20.3} & 18.07 & 16.88 \\
bo$^{\dagger}$ & 12.59 & 11.08 & 9.48 & 6.33 & 36.67 & 28.36 & \textbf{39.17} & 33.53 \\
\hdashline
Avg. & 28.72 & 25.87 & 24.88 & 24.18 & 41.42 & 36.65 & \textbf{42.2} & \underline{43.65} \\
\midrule
Total avg. & 67.88 & 67.89 & 67.27 & 67.57 & 71.71 & 70.32 & \textbf{71.92} & \underline{74.11} \\
\bottomrule
\end{tabular}
\caption{F1 scores comparison across different adapters for mBERT in ConceptNet and Glot for topic classification. All results are averaged over 3 independent runs with different random seeds.}
\label{tab:model_comparison_tc_mbert}
\end{table}

\newpage
\section{Topic Classification Results - Part II}
\label{app:tc_2}
\begin{table}[h!]
\centering
\small
\def\arraystretch{1.05}
\begin{tabular}{l|c:ccc|ccc|c}
\toprule
\multirow{2}{*}{\textbf{ISO}} & \multicolumn{8}{c}{\textbf{XLM-R}} \\
\cmidrule(lr){2-9}
 &  \multicolumn{1}{c}{} & \multicolumn{3}{c}{\textbf{ConceptNet}} & \multicolumn{4}{c}{\textbf{Glot}} \\
\cmidrule(lr){3-5} \cmidrule(lr){6-9}
& \textbf{Base} & \textbf{\texttt{Seq\_bn}} & \textbf{\texttt{LoRA}} & \textbf{\texttt{Seq\_bn\_inv}} & \textbf{\texttt{Seq\_bn}} & \textbf{\texttt{LoRA}} & \textbf{\texttt{Seq\_bn\_inv}} & \textbf{FFT} \\
\midrule
th & 87.93 & 87.19 & 87.22 & 85.99 & 86.97 & 86.8 & \textbf{88.5} & 84.21 \\
ro & 86.94 & 87.0 & 86.85 & \textbf{88.02} & 87.47 & 86.95 & 87.6 & \underline{88.03} \\
bg & 86.55 & 86.0 & \textbf{87.81} & 86.41 & 86.46 & 86.33 & 86.19 & 87.53 \\
da & 86.04 & 84.94 & 83.7 & 84.26 & 86.47 & 84.88 & \textbf{86.41} & \underline{87.06} \\
el & \textbf{86.74} & 85.59 & 85.64 & 84.32 & 85.77 & 85.28 & 86.6 & \underline{88.1} \\
he & 85.02 & 84.9 & 83.8 & 84.79 & \textbf{86.62} & 84.19 & 83.36 & 84.67 \\
sk & \textbf{87.18} & 85.53 & 84.81 & 85.2 & 85.46 & 86.52 & 86.03 & 85.59 \\
sl & 85.47 & 86.24 & \textbf{86.95} & 86.28 & 84.94 & 86.67 & 85.28 & \underline{88.12} \\
lv & 86.25 & 87.83 & 86.93 & \textbf{88.97} & 85.22 & 86.38 & 87.41 & 87.52 \\
ms & \textbf{88.12} & 87.11 & 85.82 & 85.81 & 87.94 & 85.21 & 87.94 & \underline{89.49} \\
ka & 84.08 & \textbf{85.37} & 83.79 & 83.18 & 83.92 & 85.0 & 83.95 & 82.27 \\
bn & 80.29 & 81.11 & 80.85 & 82.09 & \textbf{83.56} & 82.59 & 83.38 & \underline{84.95} \\
az & 84.05 & 85.86 & 84.24 & 85.07 & 84.43 & 85.16 & \textbf{86.39} & 86.08 \\
ur & \textbf{83.25} & 81.04 & 80.29 & 82.35 & 82.97 & 81.98 & 82.17 & \underline{83.97} \\
mk & 86.45 & 86.41 & 86.99 & 85.45 & 86.94 & 85.97 & \textbf{87.15} & \underline{88.15} \\
te & 83.58 & \textbf{83.64} & 84.26 & 83.13 & 82.43 & 84.13 & 83.43 & \underline{85.65} \\
ne & 84.14 & 83.98 & 83.92 & 83.77 & 82.65 & \textbf{84.71} & 82.85 & 84.2 \\
si & 84.92 & 84.54 & 84.86 & 82.23 & 84.49 & 83.37 & \textbf{84.99} & 84.53 \\
mr & 81.03 & 82.84 & 81.34 & 80.08 & 82.2 & 79.54 & \textbf{84.23} & 84.21 \\
sw & 77.83 & 75.58 & 76.23 & 77.97 & 80.23 & 78.73 & \textbf{81.57} & \underline{85.95} \\
cy & 79.54 & 78.44 & 80.1 & 78.99 & 78.83 & 79.15 & \textbf{81.37} & \underline{85.17} \\
am & 77.5 & 78.4 & 77.93 & 77.91 & 80.67 & 77.52 & \textbf{81.51} & \underline{84.22} \\
uz & 81.93 & 78.73 & 78.43 & 76.97 & \textbf{83.35} & 81.13 & 80.68 & \underline{86.37} \\
ku & 13.49 & 14.09 & 15.76 & 17.28 & 68.57 & 46.29 & \textbf{73.97} & \underline{81.72} \\
ug & 79.56 & 79.11 & 78.67 & 78.86 & 81.29 & \textbf{82.23} & 80.14 & \underline{84.95} \\
jv & 81.35 & 79.32 & 82.23 & 81.43 & \textbf{83.59} & 81.84 & 81.74 & 81.2 \\
su & 81.5 & 81.25 & 79.65 & 80.42 & 84.51 & 83.86 & \textbf{84.66} & 84.49 \\
\hdashline
Avg. & 81.14 & 80.82 & 80.71 & 80.64 & 83.63 & 82.31 & \textbf{84.06} & \underline{85.61} \\
\midrule
mt$^{\ddagger}$ & 64.56 & 63.62 & 61.43 & 64.43 & 77.39 & 69.74 & \textbf{77.92} & \underline{84.35} \\
bo$^{\ddagger}$ & 10.69 & 9.89 & 9.73 & 11.74 & 17.65 & \textbf{17.85} & 16.93 & \underline{20.41} \\
yo$^{\ddagger}$ & 28.29 & 26.06 & 16.07 & 24.6 & 54.13 & 35.24 & \textbf{59.44} & \underline{67.13} \\
\hdashline
Avg. & 34.52 & 33.19 & 29.08 & 33.59 & 49.72 & 40.94 & \textbf{51.43} & \underline{57.3} \\
\midrule
Total avg. & 76.48 & 76.05 & 75.54 & 75.93 & 80.24 & 78.17 & \textbf{80.79} & \underline{82.77} \\
\bottomrule
\end{tabular}
\caption{F1 scores comparison across different adapters for XLM-R in ConceptNet and Glot for topic classification. All results are averaged over 3 independent runs with different random seeds.}
\label{tab:model_comparison_tc_xlm-r}
\end{table}

\newpage
\section{Correlation Between Topic Classification and Pre- and Post-training Data}
\label{app:task_cor_1}
\begin{figure}[h]
    \centering
    \includegraphics[width=0.95\textwidth]{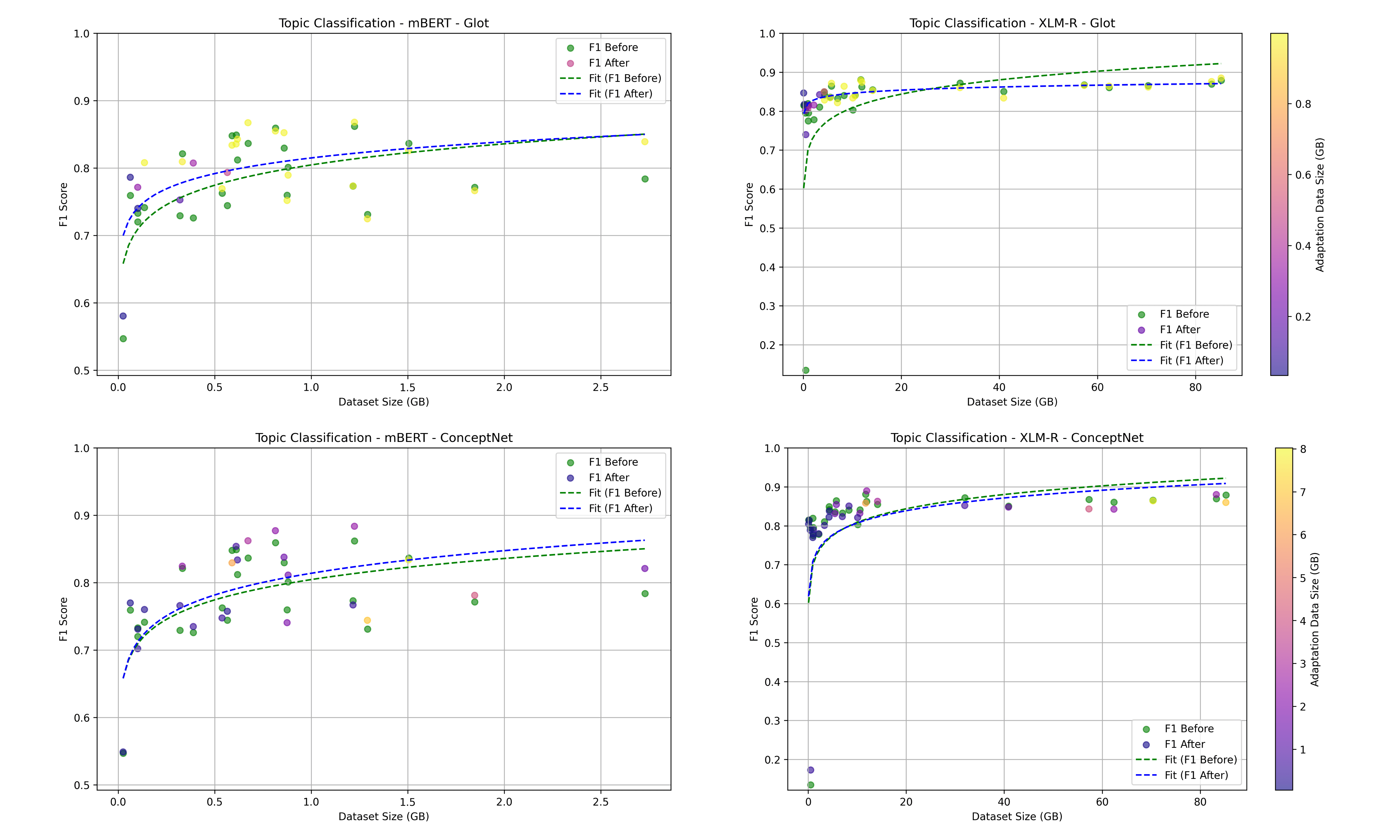}
    \caption{Correlation between the downstream performance for mBERT and XLM-R and the pre-training data and adaptation data.}
    \label{fig:tc_vs_data}
\end{figure}

\begin{table}[h]
\centering
\small
\begin{tabular}{cc|cc:cc:cc}
\toprule
\textbf{Model} & \textbf{Task} & \multicolumn{2}{c|}{\textbf{Pre-Adapt}} & \multicolumn{2}{c}{\textbf{Post-Adapt (Glot)}} & \multicolumn{2}{c}{\textbf{Post-Adapt (CN)}} \\
\cmidrule{3-8}
& & \textbf{P (p-value)} & \textbf{S (p-value)} & \textbf{P (p-value)} & \textbf{S (p-value)} & \textbf{P (p-value)} & \textbf{S (p-value)} \\
\midrule
\multirow{1}{*}{mBERT} 
& TC & 0.35 (\textit{0.1}) & 0.53 (\textit{0.008}) & 0.45 (\textit{0.03}) & 0.32 (\textit{0.13}) & 0.38 ( \textit{0.06}) & 0.55 (\textit{0.006}) \\
\midrule
\multirow{1}{*}{XLM-R} 
& TC & 0.28 (\textit{0.16}) & 0.82 (\textit{<0.005}) & 0.55 (\textit{0.002}) & 0.75 (\textit{<0.005}) & 0.28 (\textit{0.15}) & 0.83 (\textit{<0.005}) \\
\bottomrule
\end{tabular}
\caption{Pearson and Spearman Correlations for mBERT and XLM-R (Pre-Adapt and Post-Adapt) between task performance and data amounts. Post-Adapt is represented by the models adapted with the Seq\_bn\_inv language adapters and denote the correlation between the sum of the pre-training and adaptation data sizes and downstream task performance scores after the adaptation.}
\label{tab:cor_adapt_tc}
\end{table}

\newpage
\section{Named Entity Recognition Results - Part I}
\label{app:ner_1}
\begin{table}[h!]
\centering
\small
\def\arraystretch{1.05}
\begin{tabular}{l|c:ccc|ccc|cc}
\toprule
\multirow{2}{*}{\textbf{ISO}} & \multicolumn{9}{c}{\textbf{mBERT}} \\
\cmidrule(lr){2-10}
 &  \multicolumn{1}{c}{} & \multicolumn{3}{c}{\textbf{ConceptNet}} & \multicolumn{3}{c}{\textbf{Glot}} & \multicolumn{2}{c}{\textbf{Fusion}}  \\
 \cmidrule(lr){3-5} \cmidrule(lr){6-8} \cmidrule(lr){9-10}
& \textbf{Base} & \textbf{\texttt{Seq\_bn}} & \textbf{\texttt{LoRA}} & \textbf{\texttt{Seq\_bn\_inv}} & \textbf{\texttt{Seq\_bn}} & \textbf{\texttt{LoRA}} & \textbf{\texttt{Seq\_bn\_inv}} & \textbf{\texttt{Seq\_bn}} & \textbf{\texttt{Seq\_bn\_inv}} \\
 \midrule
he & 84.46 & 84.1 & 84.24 & 84.59 & 83.57 & 84.22 & 83.89 & \textbf{84.84} & 84.53  \\
el & 90.16 & 90.11 & 90.45 & 90.27 & 89.9 & \textbf{90.5} & 89.35 & 90.3 & 90.0 \\
bg & 91.25 & 91.64 & 91.64 & 91.48 & 91.64 & 91.59 & 91.56 & \textbf{91.78} & 91.76 \\
th & 67.34 & 65.65 & 66.79 & 66.68 & 67.22 & \textbf{67.8} & 66.95 & 67.36 & 67.57 \\
ro & 91.61 & 91.88 & 91.85 & 91.89 & 91.74 & 91.65 & 91.79 & 91.69 & \textbf{92.17} \\
bn & 95.46 & 96.07 & 95.82 & \textbf{96.49} & 96.42 & 96.03 & 96.3 & 95.86 & 96.1 \\
te & 75.41 & 76.17 & 76.94 & 75.29 & 75.51 & 74.69 & 75.37 & 76.53 & \textbf{77.02} \\
ka & \textbf{86.17} & 86.07 & 86.11 & 86.05 & 85.32 & 85.89 & 85.71 & 86.05 & 86.07 \\
mk & 92.43 & 92.09 & 92.3 & 92.2 & \textbf{92.62} & 92.2 & 92.02 & 91.61 & 91.98 \\
da & 89.76 & 90.08 & \textbf{90.33} & 89.74 & 90.02 & 89.72 & 88.99 & 89.41 & 89.48 \\
sl & 92.61 & 92.85 & 92.82 & 92.78 & \textbf{92.93} & 92.62 & 92.77 & 92.71 & 92.56 \\
az & 87.81 & 87.54 & 87.8 & \textbf{88.23} & 87.27 & 87.3 & 87.3 & 86.46 & 87.22 \\
sk & 90.87 & 90.88 & 90.89 & 90.96 & 90.83 & \textbf{91.3} & 90.99 & \textbf{91.04} & 90.84 \\
ms & 93.26 & 93.0 & 92.98 & 92.95 & 93.16 & \textbf{93.93} & 93.47 & 92.65 & 92.59 \\
uz & 86.48 & 86.69 & 86.58 & 86.33 & 86.87 & 86.46 & 87.73 & 87.5 & \textbf{88.45} \\
ur & 94.37 & 94.2 & 93.93 & 94.23 & 94.4 & 94.26 & 94.29 & 94.25 & \textbf{94.85} \\
cy & 88.72 & 89.34 & 89.35 & 89.05 & 89.18 & 89.36 & \textbf{90.02} & 88.95 & 88.71 \\
lv & 92.78 & 92.82 & 93.25 & 93.16 & 92.7 & 92.94 & 92.64 & \textbf{93.34} & 92.66 \\
mr & \textbf{86.34} & 86.19 & 85.97 & 86.29 & 86.32 & 86.07 & 84.35 & 86.24 & 86.22 \\
ne & 66.45 & 61.96 & 61.75 & 64.56 & \textbf{71.12} & 69.37 & 70.46 & 70.18 & 66.89 \\
jv & 52.87 & 62.83 & 61.76 & \textbf{65.3} & 63.97 & 58.73 & 63.34 & 57.21 & 58.67 \\
sw & 83.41 & 83.44 & 83.99 & 83.54 & 83.4 & 83.79 & \textbf{84.07} & 81.68 & 81.96 \\
su & 52.62 & 55.88 & 53.72 & 57.53 & 49.48 & 50.79 & 51.6 & 57.12 & \textbf{57.74} \\
yo & 79.0 & 83.02 & \textbf{83.87} & 83.1 & 81.48 & 79.58 & 79.74 & 79.81 & 78.54 \\
\hdashline
Avg. & 83.82 & 84.35 & 84.38 & \textbf{84.7} & 84.46 & 84.2 & 84.36 & 84.36 & 84.36 \\
\midrule
mt$^{\dagger}$ & 58.3 & 49.01 & 51.58 & 50.46 & 60.55 & 61.41 & \textbf{64.93} & 60.32 & 62.93 \\
ku$^{\dagger}$  & 52.34 & 60.41 & \textbf{59.92} & 59.39 & 59.9 & 52.93 & 51.51 & 52.33 & 52.4 \\
ug$^{\dagger}$ & 34.1 & 35.33 & 33.07 & 34.56 & 40.2 & 36.24 & 37.62 & 42.93 & \textbf{44.05} \\
si$^{\dagger}$ & 16.59 & 13.41 & 14.06 & 13.94 & 22.97 & 14.58 & 19.94 & 20.7 & \textbf{24.24} \\
am$^{\dagger}$ & 37.88 & 33.02 & 33.7 & 35.23 & 32.72 & 46.46 & \textbf{46.49} & 36.94 & 32.45 \\
bo$^{\dagger}$ & \textbf{56.04} & 56.02 & 55.57 & 55.29 & 53.92 & 55.45 & 53.38 & 52.03 & 53.53 \\
\hdashline
Avg. & 42.54 & 41.2 & 41.32 & 41.48 & 45.04 & 44.51 & \textbf{45.64} & 44.21 & 44.93 \\
\midrule
Total avg. & 75.56 & 75.72 & 75.77 & 76.05 & 76.58 & 76.26 & \textbf{76.62} & 76.33 & 76.47 \\
\bottomrule
\end{tabular}
\caption{F1 scores comparison for mBERT in ConceptNet and Glot for named entity recognition. All results are averaged over 3 independent runs with different random seeds.}
\label{tab:ner_mbert}
\end{table}

\newpage
\section{Named Entity Recognition Results - Part II}
\label{app:ner_2}

\begin{table}[h!]
\centering
\small
\def\arraystretch{1.05}
\begin{tabular}{l|c:ccc|ccc|cc}
\toprule
\multirow{2}{*}{\textbf{ISO}} & \multicolumn{9}{c}{\textbf{XLM-R}} \\
\cmidrule(lr){2-10}
 & & \multicolumn{3}{c}{\textbf{ConceptNet}} & \multicolumn{3}{c}{\textbf{Glot}} & \multicolumn{2}{c}{\textbf{Fusion}} \\
 \cmidrule(lr){3-5} \cmidrule(lr){6-8} \cmidrule(lr){9-10}
& \textbf{Base} & \textbf{\texttt{Seq\_bn}} & \textbf{\texttt{LoRA}} & \textbf{\texttt{Seq\_bn\_inv}} & \textbf{\texttt{Seq\_bn}} & \textbf{\texttt{LoRA}} & \textbf{\texttt{Seq\_bn\_inv}} & \textbf{\texttt{Seq\_bn}} & \textbf{\texttt{Seq\_bn\_inv}} \\
 \midrule
th & 66.55 & 66.4 & \textbf{66.85} & 66.76 & 66.63 & 65.29 & 66.2 & 65.89 & 66.82 \\
ro & 91.78 & 91.79 & 91.78 & 91.92 & 92.0 & 91.87 & \textbf{92.18} & 92.02 & 92.05 \\
bg & 91.09 & 91.22 & 91.36 & \textbf{91.48} & 91.34 & 91.4 & 91.43 & 90.91 & 91.43 \\
da & 89.58 & 89.57 & 89.54 & 89.45 & 89.44 & 89.85 & 89.72 & 89.85 & \textbf{89.89} \\
el & 90.03 & 90.32 & 89.88 & 90.14 & 89.89 & 90.02 & 90.02 & 90.18 & \textbf{90.5} \\
he & 85.56 & 85.48 & 85.45 & 84.99 & 84.92 & \textbf{85.69} & 85.28 & 85.35 & 85.4 \\
sk & 91.36 & 91.19 & 91.21 & 91.26 & 91.32 & 91.45 & \textbf{91.49} & 91.4 & 91.24 \\
sl & 92.28 & \textbf{92.58} & 92.16 & 92.41 & 92.36 & 92.05 & 92.33 & 92.21 & 92.12 \\
lv & 92.64 & 92.73 & 92.65 & 92.95 & 92.84 & 92.88 & \textbf{93.1} & 92.99 & 92.93 \\
ms & 92.0 & 92.36 & 91.65 & 92.28 & 92.4 & 92.06 & 91.9 & \textbf{92.67} & 91.82 \\
ka & 86.96 & 86.77 & 86.88 & \textbf{87.73} & 87.31 & 87.66 & 87.37 & 86.59 & 87.33 \\
bn & 95.87 & 95.66 & 95.9 & 96.06 & 96.07 & 96.13 & 96.09 & 95.57 & \textbf{96.23} \\
az & 86.13 & 85.34 & 86.47 & 86.53 & 87.03 & 86.63 & \textbf{87.59} & 86.23 & 86.38 \\
ur & 95.02 & 94.57 & \textbf{95.04} & 94.86 & 94.43 & 94.89 & 94.27 & 94.4 & 94.56 \\
mk & 92.97 & 92.47 & \textbf{93.26} & 92.28 & 92.83 & 92.68 & 92.72 & 92.32 & 92.46 \\
te & 74.67 & 73.64 & \textbf{76.07} & 74.27 & 75.18 & 74.38 & 74.82 & 72.92 & 73.91 \\
ne & 55.47 & 53.0 & 60.02 & 60.0 & 59.08 & 54.99 & 56.61 & \textbf{67.84} & 67.34 \\
si & 63.85 & 58.43 & 63.83 & 57.43 & 68.15 & 60.34 & 66.2 & 71.94 & \textbf{73.66} \\
mr & 85.92 & 85.86 & 85.5 & 85.77 & 84.75 & 85.25 & \textbf{86.1} & 85.8 & 85.52 \\
sw & 84.34 & 83.31 & 84.37 & 84.26 & \textbf{84.72} & 84.4 & 84.47 & 84.56 & 83.5 \\
cy & 89.33 & 88.9 & 88.88 & 88.97 & 89.3 & \textbf{89.72} & 89.41 & 89.4 & 89.36 \\
am & 51.22 & 49.9 & 49.29 & 48.18 & 52.57 & 47.17 & 51.67 & \textbf{55.0} & 52.55 \\
uz & 89.64 & 88.66 & 87.51 & 87.89 & 88.64 & \textbf{89.97} & 86.86 & 89.05 & 87.64 \\
ku & 35.34 & 39.53 & 42.99 & 43.83 & 40.41 & 31.43 & 29.4 & \textbf{58.02} & 56.93 \\
ug & 42.36 & 52.63 & 50.67 & 51.98 & 49.88 & 50.5 & 52.63 & 53.12 & \textbf{58.5} \\
jv & 42.99 & 45.64 & 44.7 & 50.87 & 46.51 & 44.7 & 47.96 & \textbf{63.53} & 58.81 \\
su & 33.07 & 38.4 & 42.26 & 48.32 & 41.47 & 39.76 & 42.89 & \textbf{52.53} & 49.61 \\
\hdashline
Avg. & 77.33 & 77.64 & 78.38 & 78.62 & 78.57 & 77.52 & 78.17 & \textbf{80.83} & 80.68 \\
\midrule
mt$^{\ddagger}$  & 46.31 & 32.69 & 40.11 & 32.13 & 48.03 & 41.54 & 53.57 & \textbf{64.31} & 57.57 \\
bo$^{\ddagger}$  & 43.51 & 44.29 & 44.55 & 46.41 & 41.86 & 39.64 & 38.27 & \textbf{48.15} & 47.55 \\
yo$^{\ddagger}$  & 73.54 & 71.2 & 73.46 & 74.59 & 73.3 & 74.87 & 75.09 & 73.04 & \textbf{75.8} \\
\hdashline
Avg. & 54.45 & 49.39 & 52.71 & 51.04 & 54.4 & 52.01 & 55.64 & \textbf{61.83} & 60.31 \\
\midrule
Total avg. & 75.05 & 74.82 & 75.81 & 75.87 & 76.16 & 74.97 & 75.92 & \textbf{78.93} & 78.65 \\
\bottomrule
\end{tabular}
\caption{F1 scores for XLM-R across ConceptNet and Glot for named entity recognition. All results are averaged over 3 independent runs with different random seeds.}
\label{tab:ner_xlm-r}
\end{table}

\newpage
\section{Correlation Between Named Entity Recognition and Pre- and Post-training data}
\label{app:task_cor_2}
\begin{figure}[h]
    \centering
    \includegraphics[width=0.95\textwidth]{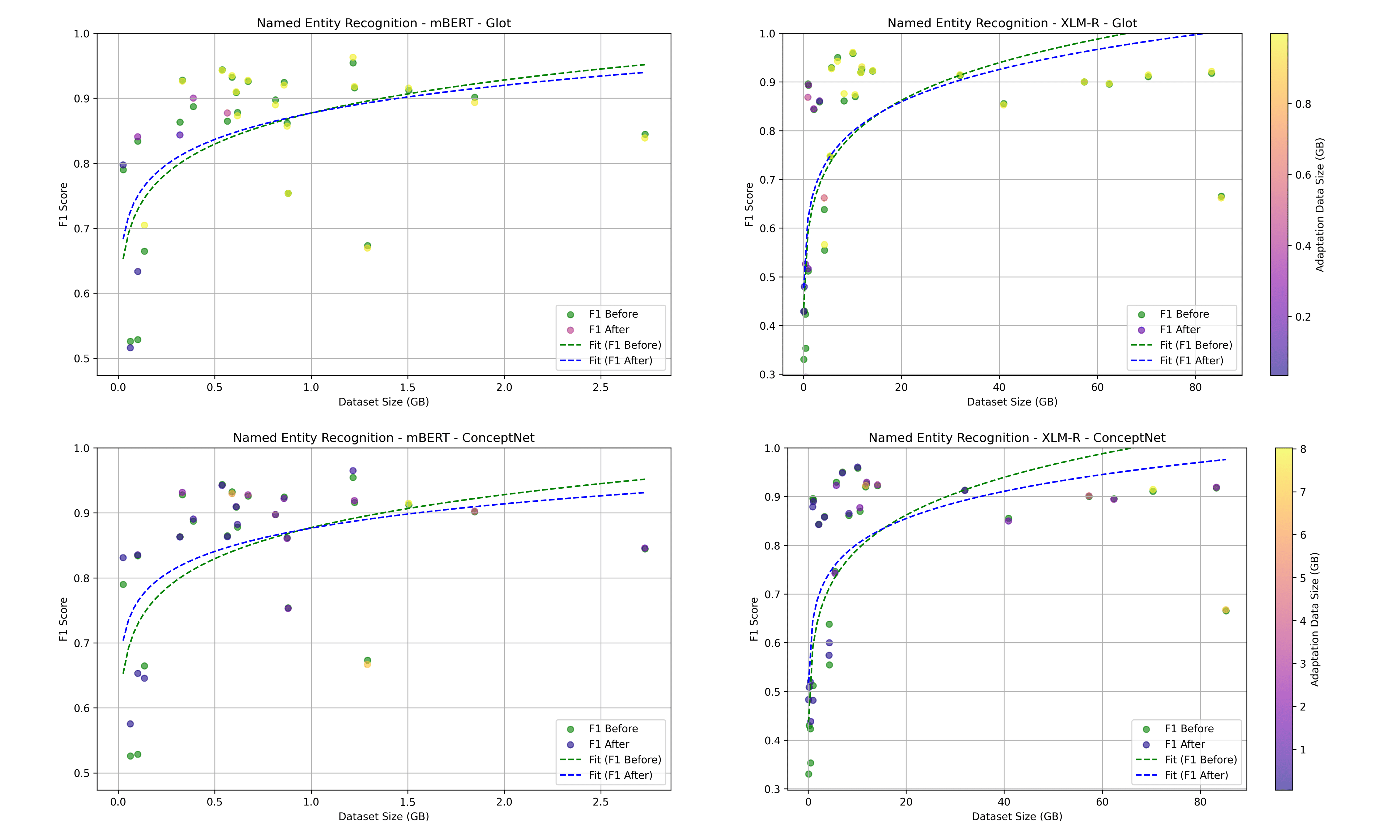}
    \caption{Correlation between the downstream performance for mBERT and XLM-R and the pre-training data and adaptation data.}
    \label{fig:ner_vs_data}
\end{figure}

\begin{table}[h]
\centering
\small
\begin{tabular}{cc|cc:cc:cc}
\toprule
\textbf{Model} & \textbf{Task} & \multicolumn{2}{c|}{\textbf{Pre-Adapt}} & \multicolumn{2}{c}{\textbf{Post-Adapt (Glot)}} & \multicolumn{2}{c}{\textbf{Post-Adapt (CN)}} \\
\cmidrule{3-8}
& & \textbf{P (p-value)} & \textbf{S (p-value)} & \textbf{P (p-value)} & \textbf{S (p-value)} & \textbf{P (p-value)} & \textbf{S (p-value)} \\
\midrule
\multirow{1}{*}{mBERT} 
& NER & 0.32 (\textit{0.1}) & 0.32 (\textit{0.1}) & 0.42 (\textit{0.04}) & 0.29 (\textit{0.2}) & 0.20 ( \textit{0.3}) & 0.44 (\textit{0.03}) \\
\midrule
\multirow{1}{*}{XLM-R} 
& NER & 0.31 (\textit{0.1}) & 0.58 (\textit{0.002}) & 0.31 (\textit{0.1}) & 0.61 (\textit{<0.005}) & 0.32 (\textit{0.1}) & 0.60 (\textit{<0.005}) \\
\bottomrule
\end{tabular}
\caption{Pearson and Spearman Correlations for mBERT and XLM-R (Pre-Adapt and Post-Adapt) between task performance and data amounts. Post-Adapt is represented by the models adapted with the Seq\_bn\_inv language adapters and denote the correlation between the sum of the pre-training and adaptation data sizes and downstream task performance scores after the adaptation.}
\label{tab:cor_adapt_ner}
\end{table}

\newpage
\section{Sentiment Analysis Results - Part I}
\label{app:sa_1}
\begin{table}[h!]
\centering
\small
\def\arraystretch{1.05}
\begin{tabular}{l|c:ccc|ccc|c}
\toprule
\multirow{2}{*}{\textbf{ISO}} & \multicolumn{8}{c}{\textbf{mBERT}} \\
\cmidrule(lr){2-9}
 &  \multicolumn{1}{c}{} & \multicolumn{3}{c}{\textbf{ConceptNet}} & \multicolumn{4}{c}{\textbf{Glot}} \\
\cmidrule(lr){3-5} \cmidrule(lr){6-9}
& \textbf{Base} & \textbf{\texttt{Seq\_bn}} & \textbf{\texttt{LoRA}} & \textbf{\texttt{Seq\_bn\_inv}} & \textbf{\texttt{Seq\_bn}} & \textbf{\texttt{LoRA}} & \textbf{\texttt{Seq\_bn\_inv}} & \textbf{FFT} \\
\midrule
he & 91.42 & 91.55 & 90.44 & 90.81 & 90.79 & 90.87 & \textbf{91.58} & 90.6 \\
el & \textbf{86.35} & 86.27 & 86.05 & 86.22 & 84.88 & 84.95 & 84.52 & \underline{86.38} \\
bg & 88.82 & 89.41 & 89.17 & \textbf{89.54} & 88.76 & 88.65 & 89.2 & \underline{89.99} \\
th & 81.68 & 81.97 & 81.92 & 82.45 & 82.57 & 82.0 & \textbf{83.23} & \underline{83.19} \\
ro & 92.87 & 92.67 & 92.62 & 92.64 & 93.13 & \textbf{92.98} & 92.96 & \underline{93.7} \\
bn & 92.28 & 92.16 & 92.6 & 91.88 & 92.26 & 92.56 & \textbf{92.57} & \underline{92.48} \\
te & 83.49 & 83.29 & 84.17 & 85.01 & \textbf{85.55} & 84.45 & 85.26 & \underline{88.41} \\
ka & 78.12 & 78.1 & 76.68 & 76.05 & 80.03 & 80.23 & \textbf{81.24} & \underline{86.97} \\
mk & 62.47 & \textbf{69.01} & 66.4 & 62.07 & 67.54 & 65.06 & 65.21 & 68.98 \\
da & 95.71 & 95.33 & 95.77 & 95.33 & 95.95 & \textbf{96.15} & 96.09 & \underline{96.84} \\
sl & 85.71 & 86.46 & 86.28 & 85.81 & 86.79 & 86.4 & \textbf{87.83} & \underline{88.66} \\
az & 79.42 & 79.59 & 79.72 & 80.03 & 79.62 & \textbf{80.15} & 80.13 & \underline{81.44} \\
sk & 91.11 & 88.86 & 89.9 & 89.73 & 90.87 & 91.16 & \textbf{92.18} & 91.08 \\
ms & 91.5 & 92.03 & 91.87 & 91.99 & 92.06 & 91.7 & \textbf{92.57} & \underline{93.83} \\
uz & 86.84 & 85.67 & 86.76 & 85.85 & 86.52 & 86.36 & \textbf{86.85} & \underline{88.33} \\
ur & 82.43 & 81.89 & 82.01 & 82.13 & 82.69 & 82.66 & \textbf{82.72} & \underline{83.81} \\
cy & 87.28 & 86.99 & 87.82 & 86.15 & 87.71 & \textbf{87.76} & 87.42 & \underline{88.53} \\
lv & 75.41 & 75.66 & 73.99 & 74.71 & 76.32 & 75.41 & \textbf{76.65} & \underline{79.24} \\
mr & 88.7 & 88.76 & 89.0 & 88.67 & \textbf{89.43} & 89.13 & 88.97 & \underline{90.43} \\
ne & 59.51 & 51.46 & \textbf{67.17} & 55.31 & 56.77 & 59.35 & 63.19 & \underline{63.47} \\
jv & 75.38 & 74.24 & 74.75 & 73.94 & \textbf{76.16} & 75.7 & 75.43 & 75.44 \\
sw & 57.71 & 54.25 & 57.24 & 52.9 & 65.05 & 62.21 & \textbf{69.64} & 54.6 \\
su & 82.13 & 84.25 & 84.62 & 83.33 & 84.42 & \textbf{84.75} & 83.99 & 84.06 \\
yo & 76.1 & 75.66 & 75.24 & 75.35 & 75.93 & 75.43 & \textbf{77.85} & 77.32 \\
\hdashline
Avg. & 82.18 & 81.9 & 82.59 & 81.58 & 82.99 & 82.75 & \textbf{83.64} & \underline{84.07} \\
\midrule
mt$^{\dagger}$ & 65.24 & 65.68 & 62.82 & 66.88 & 68.79 & \textbf{73.87} & 65.34 & \underline{74.11} \\
ku$^{\dagger}$ & 84.2 & 82.82 & 83.97 & 83.37 & 85.14 & 84.46 & \textbf{86.14} & 85.55 \\
ug$^{\dagger}$ & 70.94 & 68.35 & 72.67 & 72.19 & 76.91 & 71.35 & \textbf{80.4} & 76.63 \\
si$^{\dagger}$ & 64.97 & 64.89 & 65.01 & 64.67 & 65.42 & \textbf{66.02} & 65.62 & \underline{66.26} \\
am$^{\dagger}$ & 61.45 & 62.02 & 60.87 & 61.45 & 60.3 & 61.62 & \textbf{63.81} & 59.48 \\
bo$^{\dagger}$ & 79.4 & 79.12 & 79.38 & 80.67 & \textbf{83.27} & 82.33 & 82.14 & 81.77 \\
\hdashline
Avg. & 71.03 & 70.48 & 70.79 & 71.54 & 73.3 & 73.27 & \textbf{73.91} & \underline{73.97} \\
\midrule
Total avg. & 79.95 & 79.61 & 80.23 & 79.57 & 81.05 & 80.86 & \textbf{81.69} & \underline{82.05} \\
\bottomrule
\end{tabular}
\caption{F1 scores comparison across different adapters for mBERT in ConceptNet and Glot for sentiment analysis. All results are averaged over 3 independent runs with different random seeds.}
\label{tab:sa_mbert}
\end{table}

\newpage
\section{Sentiment Analysis Results - Part II}
\label{app:sa_2}
\begin{table}[h!]
\centering
\small
\def\arraystretch{1.05}
\begin{tabular}{l|c:ccc|ccc|c}
\toprule
\multirow{2}{*}{\textbf{ISO}} & \multicolumn{8}{c}{\textbf{XLM-R}} \\
\cmidrule(lr){2-9}
 &  \multicolumn{1}{c}{} & \multicolumn{3}{c}{\textbf{ConceptNet}} & \multicolumn{4}{c}{\textbf{Glot}} \\
\cmidrule(lr){3-5} \cmidrule(lr){6-9}
& \textbf{Base} & \textbf{\texttt{Seq\_bn}} & \textbf{\texttt{LoRA}} & \textbf{\texttt{Seq\_bn\_inv}} & \textbf{\texttt{Seq\_bn}} & \textbf{\texttt{LoRA}} & \textbf{\texttt{Seq\_bn\_inv}} & \textbf{FFT} \\
\midrule
th & 88.18 & 88.26 & 88.43 & \textbf{88.46} & 88.11 & 88.31 & 88.13 & 86.39 \\
ro & 94.37 & 94.84 & 95.03 & \textbf{95.04} & 94.74 & 94.67 & 95.03 & 94.55 \\
bg & 91.36 & 90.66 & \textbf{91.43} & 91.41 & 91.26 & 90.93 & 90.65 & 90.79 \\
da & 98.04 & 97.84 & \textbf{98.13} & 98.02 & 98.09 & 98.04 & 97.98 & 97.82 \\
el & 88.82 & 88.92 & \textbf{88.98} & 88.73 & 88.19 & 88.25 & 88.61 & 88.75 \\
he & 91.26 & 89.66 & \textbf{91.81} & 91.25 & 90.48 & 90.27 & 90.85 & 90.2 \\
sk & \textbf{94.6} & 93.86 & 93.87 & 93.43 & 93.22 & 93.72 & 93.44 & 94.03 \\
sl & 93.75 & 93.46 & \textbf{94.32} & 92.68 & 94.23 & 93.57 & 93.86 & 92.73 \\
lv & 83.3 & \textbf{83.78} & 83.36 & 83.83 & 82.47 & 83.12 & 83.65 & 82.97 \\
ms & 95.51 & 95.27 & \textbf{95.66} & 95.57 & 95.44 & 95.29 & 95.53 & 95.26 \\
ka & \textbf{91.92} & 91.51 & 90.8 & 91.21 & \textbf{91.92} & 91.11 & 91.41 & \underline{93.33} \\
bn & 93.78 & 94.14 & 94.3 & \textbf{94.46} & 94.13 & 94.1 & 94.43 & 94.41 \\
az & 84.05 & 84.05 & 84.05 & 83.98 & 84.32 & 84.2 & \textbf{84.74} & \underline{85.19} \\
ur & 85.6 & 85.99 & 85.67 & 85.85 & 85.89 & \textbf{86.7} & 86.25 & \underline{87.27} \\
mk & 70.96 & 69.22 & 67.05 & 69.45 & \textbf{73.9} & 70.74 & 72.31 & 71.68 \\
te & 89.72 & 89.15 & 89.59 & 89.22 & 89.56 & 89.72 & \textbf{89.9} & \underline{90.92} \\
ne & 64.6 & \textbf{69.37} & 64.06 & 63.02 & 67.49 & 68.38 & 68.65 & 65.46 \\
si & 92.49 & 92.59 & 92.18 & \textbf{93.21} & 92.49 & 91.78 & 91.96 & 92.85 \\
mr & 91.17 & 91.8 & 91.9 & 91.8 & 91.87 & \textbf{92.36} & 91.8 & \underline{92.43} \\
sw & 70.08 & 65.37 & 77.11 & 75.3 & \textbf{79.52} & 77.24 & 74.45 & \underline{83.84} \\
cy & 90.83 & 91.01 & 90.57 & 90.65 & 91.12 & 90.88 & \textbf{91.36} & 91.01 \\
am & 86.15 & 83.77 & 84.2 & 82.88 & 87.04 & \textbf{87.9} & 87.7 & 87.49 \\
uz & 87.63 & 88.24 & 88.37 & 88.13 & \textbf{88.47} & 87.98 & 88.39 & \underline{90.08} \\
ku & 89.39 & 89.73 & 89.08 & 89.78 & 92.57 & 89.09 & \textbf{93.31} & \underline{95.31} \\
ug & 88.97 & 88.88 & 89.91 & 87.64 & 88.81 & \textbf{90.01} & 89.65 & \underline{91.72} \\
jv & 76.51 & 77.34 & 77.01 & 77.14 & 76.51 & 76.79 & \textbf{77.65} & 75.53 \\
su & 88.15 & 82.66 & 85.17 & 84.41 & 89.69 & \textbf{90.34} & 89.69 & 89.03 \\
\hdashline
Avg. & 87.45 & 87.09 & 87.48 & 87.28 & \textbf{88.2} & 87.98 & \textbf{88.2} & \underline{88.56} \\
\midrule
mt$^{\ddagger}$ & 55.63 & 55.19 & 55.32 & 54.13 & \textbf{69.4} & 63.15 & 69.31 &  \underline{70.38} \\
bo$^{\ddagger}$ & 51.81 & 47.33 & 51.07 & 49.34 & \textbf{52.92} & 50.9 & 50.69 & \underline{55.19} \\
yo$^{\ddagger}$ & 74.73 & 73.4 & 73.6 & 75.09 & 75.5 & 72.0 & \textbf{77.65} & \underline{78.99} \\
\hdashline
Avg. & 60.72 & 58.64 & 60.00 & 59.52 & \textbf{65.94} & 62.02 & 65.88 & \underline{68.19} \\
\midrule
Total avg. & 84.78 & 84.24 & 84.73 & 84.50 & \textbf{85.98} & 85.38 & 85.97 & \underline{86.52} \\
\bottomrule
\end{tabular}
\caption{F1 scores comparison across different adapters for XLM-R in ConceptNet and Glot for sentiment analysis. All results are averaged over 3 independent runs with different random seeds.}
\label{tab:sa_xlmr}
\end{table}

\newpage
\section{Correlation Between Sentiment Analysis and Pre- and Post-training data}
\label{app:task_cor_3}
\begin{figure}[h]
    \centering
    \includegraphics[width=0.95\textwidth]{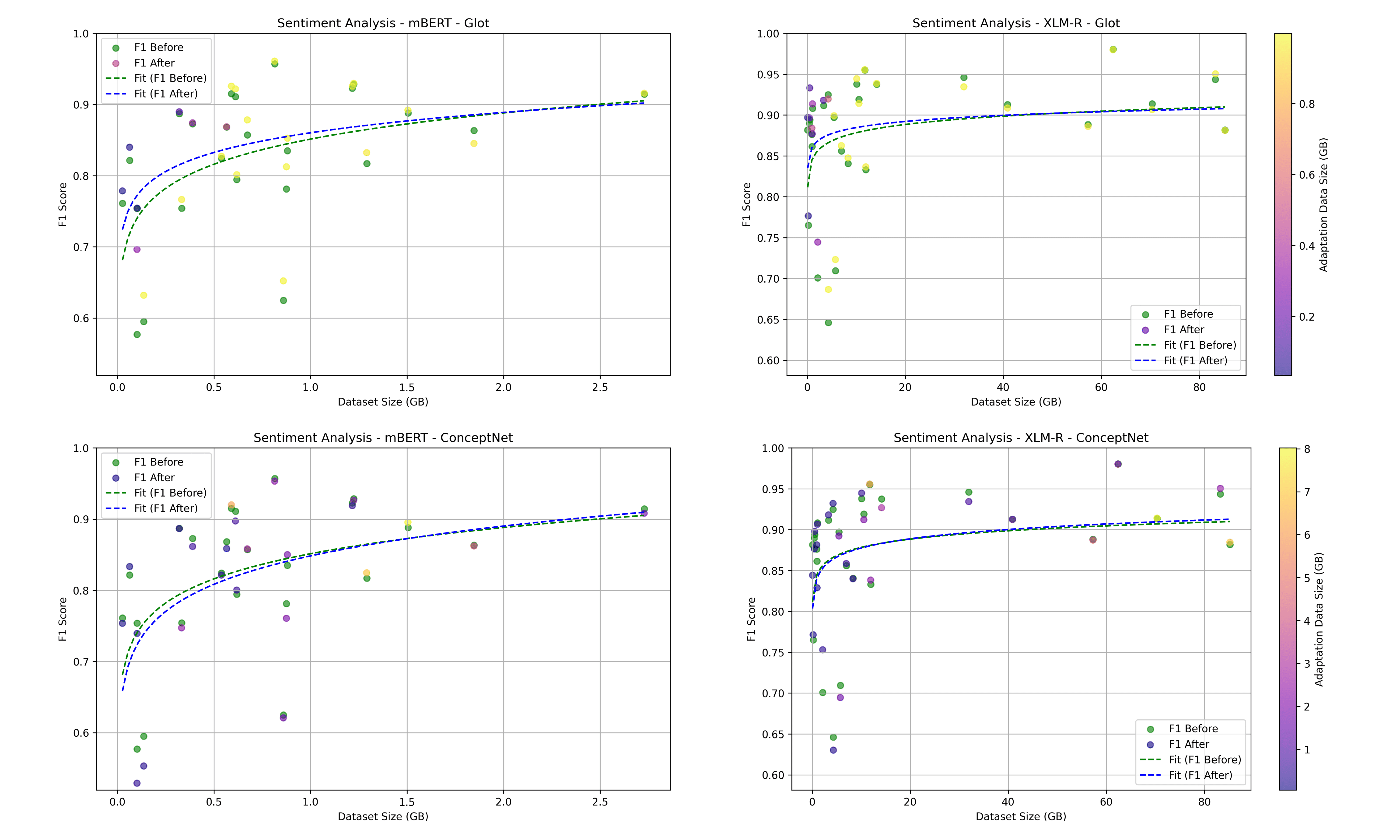}
    \caption{\centering Correlation between the downstream performance for mBERT and XLM-R and the pre-training data and adaptation data.}
    \label{fig:sa_vs_data}
\end{figure}

\begin{table}[h]
\centering
\small
\begin{tabular}{cc|cc:cc:cc}
\toprule
\textbf{Model} & \textbf{Task} & \multicolumn{2}{c|}{\textbf{Pre-Adapt}} & \multicolumn{2}{c}{\textbf{Post-Adapt (Glot)}} & \multicolumn{2}{c}{\textbf{Post-Adapt (CN)}} \\
\cmidrule{3-8}
& & \textbf{P (p-value)} & \textbf{S (p-value)} & \textbf{P (p-value)} & \textbf{S (p-value)} & \textbf{P (p-value)} & \textbf{S (p-value)} \\
\midrule
\multirow{1}{*}{mBERT} 
& SA & 0.45 (\textit{0.03}) & 0.50 (\textit{0.01}) & 0.38 (\textit{0.07}) & 0.41 (\textit{0.05}) & 0.39 ( \textit{0.06}) & 0.52 (\textit{0.009}) \\
\midrule
\multirow{1}{*}{XLM-R} 
& SA & 0.36 (\textit{0.07}) & 0.47 (\textit{0.01}) & 0.32 (\textit{0.1}) & 0.33 (\textit{0.1}) & 0.38 (\textit{0.05}) & 0.52 (\textit{0.005}) \\
\bottomrule
\end{tabular}
\caption{Pearson and Spearman Correlations for mBERT and XLM-R (Pre-Adapt and Post-Adapt) between task performance and data amounts. Post-Adapt is represented by the models adapted with the Seq\_bn\_inv language adapters and denote the correlation between the sum of the pre-training and adaptation data sizes and downstream task performance scores after the adaptation.}
\label{tab:cor_adapt_sa}
\end{table}

\newpage
\section{Results for Large-Scale Models for TC and NER}
\label{ap:prompting}

\begin{table}[h!]
\resizebox{\columnwidth}{!}{
\begin{tabular}{lccccccccccccccccc}
\toprule
 & \rotatebox{90}{GPT-3.5-turbo-0613} & \rotatebox{90}{GPT-4-0613} & \rotatebox{90}{LLaMAX2-7B-Alpaca} & \rotatebox{90}{Llama-2-7b-chat-hf} & \rotatebox{90}{Meta-Llama-3-8B} & \rotatebox{90}{Meta-Llama-3.1-8B} & \rotatebox{90}{Qwen1.5-7B} & \rotatebox{90}{Qwen2-7B} & \rotatebox{90}{bloom-7b1} & \rotatebox{90}{bloomz-7b1} & \rotatebox{90}{gemma-2-9b} & \rotatebox{90}{gemma-7b} & \rotatebox{90}{mala-500-10b-v1} & \rotatebox{90}{mala-500-10b-v2} & \rotatebox{90}{occiglot-7b-eu5} & \rotatebox{90}{xglm-7.5B} & \rotatebox{90}{yayi-7b-llama2} \\
\midrule
am & 24.14 & 38.74 & 7.64 & 5.41 & 38.03 & 40.43 & 13.57 & 23.68 & 7.32 & 8.27 & 41.19 & 43.02 & 5.71 & 9.03 & 6.85 & 7.86 & 3.59 \\
az & 52.17 & 44.27 & 30.81 & 20.54 & 73.78 & 71.97 & 51.86 & 65.86 & 10.08 & 16.68 & 57.95 & 68.79 & 5.71 & 5.71 & 31.37 & 26.56 & 17.55 \\
bn & 54.29 & 50.55 & 23.79 & 9.35 & 65.89 & 63.43 & 42.62 & 66.08 & 10.75 & 20.93 & 51.22 & 66.91 & 5.71 & 5.69 & 22.42 & 28.25 & 12.99 \\
bo & 2.90 & 1.94 & 3.69 & 4.63 & 40.80 & 48.83 & 10.15 & 12.41 & 6.44 & 10.56 & 12.12 & 20.23 & 5.71 & 3.63 & 11.65 & 7.06 & 6.61 \\
bg & 54.80 & 58.33 & 31.47 & 29.92 & 64.95 & 63.53 & 55.15 & 77.06 & 20.17 & 16.58 & 51.85 & 63.26 & 5.71 & 5.23 & 44.70 & 41.81 & 24.77 \\
ku & 38.74 & 38.10 & 19.71 & 7.67 & 68.26 & 65.47 & 21.71 & 33.20 & 10.26 & 8.63 & 33.49 & 44.59 & 5.71 & 6.86 & 14.07 & 9.31 & 7.81 \\
cy & 43.08 & 42.47 & 26.76 & 18.08 & 68.75 & 68.69 & 37.47 & 49.93 & 10.45 & 18.09 & 50.38 & 56.87 & 5.71 & 5.71 & 26.21 & 17.57 & 19.37 \\
da & 52.71 & 52.17 & 33.17 & 34.03 & 73.03 & 73.73 & 57.73 & 75.95 & 17.85 & 21.90 & 45.05 & 71.14 & 5.71 & 5.39 & 49.20 & 56.88 & 32.02 \\
el & 54.29 & 60.27 & 21.84 & 21.69 & 70.22 & 73.70 & 46.99 & 63.73 & 11.97 & 11.90 & 39.08 & 67.20 & 5.71 & 5.71 & 31.48 & 55.80 & 20.84 \\
he & 56.84 & 51.09 & 24.39 & 17.55 & 69.01 & 69.80 & 46.93 & 70.07 & 10.87 & 8.53 & 44.35 & 64.03 & 5.71 & 4.76 & 22.82 & 10.66 & 9.51 \\
jv & 21.05 & 21.05 & 28.49 & 21.31 & 66.73 & 69.39 & 49.99 & 50.76 & 17.90 & 25.19 & 59.48 & 57.33 & 5.71 & 2.20 & 34.05 & 44.85 & 19.65 \\
ka & 47.19 & 43.68 & 18.37 & 15.25 & 68.58 & 63.50 & 32.76 & 52.02 & 3.50 & 14.76 & 58.73 & 69.17 & 5.71 & 8.13 & 25.17 & 9.35 & 13.24 \\
lv & 54.29 & 53.76 & 31.62 & 23.85 & 69.79 & 70.63 & 55.05 & 67.69 & 12.70 & 17.38 & 45.97 & 69.24 & 8.21 & 3.13 & 34.20 & 23.25 & 23.91 \\
mr & 52.71 & 51.09 & 19.90 & 14.04 & 64.84 & 63.07 & 39.41 & 56.66 & 26.78 & 29.62 & 27.30 & 56.58 & 5.71 & 5.83 & 19.63 & 23.24 & 9.59 \\
mk & 52.71 & 60.75 & 28.98 & 26.75 & 66.66 & 68.33 & 55.99 & 75.87 & 12.91 & 16.69 & 55.62 & 64.88 & 3.97 & 3.49 & 40.23 & 40.43 & 22.65 \\
mt & 44.27 & 50.55 & 29.18 & 23.07 & 63.25 & 67.22 & 44.26 & 56.10 & 11.45 & 20.18 & 43.93 & 62.54 & 5.71 & 5.71 & 34.33 & 28.45 & 24.09 \\
ne & 55.83 & 52.71 & 21.49 & 18.42 & 62.32 & 62.69 & 42.96 & 54.99 & 10.12 & 19.45 & 15.71 & 62.31 & 5.62 & 4.07 & 25.79 & 31.47 & 18.61 \\
ro & 51.64 & 54.80 & 34.88 & 31.49 & 70.19 & 72.20 & 56.43 & 74.64 & 20.10 & 20.76 & 52.51 & 69.08 & 5.71 & 5.71 & 47.32 & 43.15 & 30.92 \\
si & 23.38 & 62.63 & 8.66 & 4.81 & 60.25 & 57.45 & 12.49 & 29.29 & 5.98 & 9.38 & 46.12 & 65.92 & 6.60 & 2.20 & 10.82 & 5.48 & 5.71 \\
sk & 52.17 & 52.71 & 28.65 & 29.75 & 70.57 & 72.77 & 55.40 & 74.63 & 20.27 & 17.58 & 35.52 & 68.94 & 5.71 & 8.49 & 43.66 & 39.12 & 27.70 \\
sl & 53.76 & 47.76 & 33.60 & 31.05 & 75.67 & 70.18 & 55.53 & 63.56 & 11.10 & 17.18 & 48.42 & 67.87 & 9.22 & 3.30 & 40.19 & 30.21 & 28.09 \\
su & 26.38 & 20.26 & 28.22 & 23.89 & 63.50 & 67.46 & 46.31 & 58.94 & 17.55 & 21.68 & 60.68 & 65.78 & 5.71 & 7.69 & 32.18 & 44.52 & 21.59 \\
sw & 55.83 & 46.62 & 28.24 & 14.01 & 68.95 & 68.37 & 40.51 & 51.05 & 12.91 & 22.41 & 48.61 & 58.78 & 5.71 & 6.70 & 29.03 & 45.91 & 11.88 \\
te & 57.84 & 50.00 & 5.92 & 5.78 & 64.72 & 62.36 & 27.29 & 55.69 & 16.89 & 20.13 & 47.24 & 68.93 & 5.71 & 5.17 & 12.91 & 49.59 & 4.73 \\
th & 53.24 & 49.45 & 16.94 & 20.88 & 77.50 & 75.40 & 46.57 & 67.38 & 6.25 & 16.62 & 45.24 & 58.64 & 5.71 & 7.82 & 35.27 & 50.01 & 21.98 \\
ug & 44.27 & 46.04 & 6.53 & 6.90 & 66.23 & 62.22 & 12.37 & 54.64 & 9.29 & 11.72 & 33.74 & 45.77 & 7.54 & 3.66 & 16.20 & 7.76 & 7.13 \\
ur & 53.24 & 65.79 & 22.87 & 15.07 & 67.80 & 67.53 & 39.13 & 61.90 & 23.50 & 23.33 & 29.04 & 56.48 & 5.71 & 6.61 & 29.62 & 41.90 & 12.23 \\
uz & 44.87 & 34.82 & 29.50 & 13.49 & 69.53 & 68.35 & 33.53 & 54.55 & 10.05 & 13.89 & 56.50 & 65.44 & 5.71 & 11.34 & 26.86 & 15.93 & 10.11 \\
yo & 22.61 & 16.22 & 18.17 & 11.26 & 50.05 & 46.74 & 25.36 & 30.44 & 14.08 & 21.90 & 35.16 & 37.11 & 8.75 & 7.65 & 18.71 & 16.97 & 10.23 \\
ms & 49.45 & 55.83 & 30.46 & 27.35 & 74.10 & 73.28 & 56.74 & 76.05 & 11.13 & 23.31 & 55.96 & 69.52 & 5.71 & 5.71 & 40.15 & 46.19 & 27.33 \\
\midrule
Total avg. & 45.02 & 45.82 & 23.13 & 18.24 & 65.80 & 65.62 & 40.41 & 56.83 & 13.02 & 17.51 & 44.27 & 60.21 & 6.04 & 5.74 & 28.57 & 29.98 & 16.88 \\
\bottomrule
\end{tabular}
}
\caption{F1 Scores for All Large-Scale Models on TC. The results are based on 3-shot prompting, as reported by \citet{ji2024emma500enhancingmassivelymultilingual}. GPT-3.5 and GPT-4 results are zero-shot, obtained from \citet{adelani-etal-2024-sib}.}
\label{tab:llms_tc}
\end{table}

\begin{table}[h!]
\small
\centering
\begin{tabular}{lcccccccc}
\toprule
 & Bloom & Bloomz & mT0 & GPT-3.5-turbo-0301 \\
\midrule
th & 1.0 & 0.2 & 1.4 & - \\
el & 19.7 & 13.0 & 12.8 & 69.3 \\
ur & 71.7 & 47.3 & 47.1 & - \\
te & 5.3 & 3.8 & 3.3 & - \\
sw & 58.8 & 26.8 & 24.3 & - \\
bg & 29.6 & 19.7 & 14.7 & 72.0 \\
mr & 27.9 & 20.4 & 12.3 & - \\
bn & 36.8 & 36.2 & 23.9 & - \\
\midrule
Total avg. & 31.35 & 20.92 & 17.48 & 70.65 \\
\bottomrule
\end{tabular}
\caption{Three-shot NER results across eight overlapping languages from BUFFET \cite{asai2023buffet}. The scores for GPT-3.5 are only provided for two languages.}
\label{tab:llms_ner}
\end{table}

\begin{table}[th!]
\small
\centering
\begin{tabular}{lccccc}
\toprule
& Qwen 1.5B & Qwen 7B & Llama 8B & Qwen 14B & Llama 70B \\
\midrule
am & 7.03 & 13.99 & 9.18 & 31.76 & 43.41 \\
az & 9.60 & 18.48 & 12.27 & 53.05 & 73.19 \\
be & 6.59 & 31.51 & 20.25 & 68.20 & 78.17 \\
bo & 2.38 & 8.17 & 9.67 & 18.92 & 62.63 \\
bg & 7.93 & 26.47 & 24.31 & 46.81 & 78.65 \\
ku & 6.77 & 18.48 & 20.10 & 17.98 & 77.52 \\
cy & 8.93 & 18.49 & 20.32 & 26.68 & 61.55 \\
da & 13.04 & 25.62 & 17.90 & 41.18 & 78.29 \\
el & 3.91 & 10.26 & 16.41 & 58.39 & 77.90 \\
he & 5.50 & 23.03 & 20.77 & 46.25 & 76.66 \\
jv & 10.51 & 19.45 & 19.53 & 28.04 & 66.43 \\
ka & 4.35 & 20.46 & 24.99 & 45.74 & 77.60 \\
lv & 11.14 & 14.29 & 17.60 & 44.14 & 74.09 \\
mr & 6.17 & 22.67 & 22.31 & 49.54 & 68.77 \\
mk & 4.91 & 24.16 & 22.44 & 44.38 & 77.66 \\
mt & 11.76 & 18.01 & 18.24 & 49.23 & 66.83 \\
ne & 4.70 & 23.59 & 26.36 & 55.34 & 69.25 \\
ro & 9.50 & 21.93 & 24.67 & 57.25 & 77.72 \\
si & 12.47 & 14.28 & 14.96 & 29.43 & 70.69 \\
sk & 6.66 & 15.61 & 21.37 & 45.38 & 75.80 \\
sl & 13.34 & 22.71 & 18.89 & 43.22 & 65.42 \\
su & 9.44 & 22.41 & 21.98 & 34.95 & 65.53 \\
sw & 10.38 & 11.15 & 15.45 & 18.60 & 67.94 \\
te & 9.19 & 17.90 & 27.21 & 38.99 & 75.35 \\
th & 8.49 & 40.22 & 20.80 & 73.49 & 74.23 \\
ug & 7.02 & 17.72 & 19.67 & 28.83 & 71.21 \\
ur & 3.71 & 27.47 & 24.23 & 47.75 & 80.07 \\
uz & 11.76 & 21.45 & 17.02 & 38.32 & 70.58 \\
yo & 6.70 & 13.49 & 15.20 & 18.57 & 45.55 \\
ms & 10.58 & 21.73 & 27.82 & 56.00 & 73.02 \\
\midrule
Total avg. & 8.15 & 20.17 & 19.73 & 41.88 & 70.72 \\
\bottomrule
\end{tabular}
\caption{F1 Scores for DeepSeek-R1 distilled models of various sizes for TC. The results are based on zero-shot prompting and were obtained in our evaluation.}
\label{tab:f1_scores_avg}
\end{table}

\begin{table}[th!]
\small
\centering
\begin{tabular}{lcccccc}
\toprule
\textbf{Language} & \multicolumn{2}{c}{\textbf{TC}} & \multicolumn{2}{c}{\textbf{NER}} & \multicolumn{2}{c}{\textbf{SA}} \\
\cmidrule(lr){2-3} \cmidrule(lr){4-5} \cmidrule(lr){6-7}
 & \textbf{\makecell{LLaMA-3 \\ (Baseline)}} & \textbf{\makecell{LLaMA-3\\+\texttt{Seq\_bn\_inv}}} & \textbf{\makecell{LLaMA-3 \\ (Baseline)}} & \textbf{\makecell{LLaMA-3\\+\texttt{Seq\_bn\_inv}}} & \textbf{\makecell{LLaMA-3 \\ (Baseline)}} & \textbf{\makecell{LLaMA-3\\+\texttt{Seq\_bn\_inv}}} \\
\midrule
cy & 33.64 & 72.50 & 76.36 & 77.03 & 58.36 & 88.43 \\
si & 16.67 & 39.11 & 30.84 & 30.08 & 80.42 & 83.8 \\
sw & 29.05 & 60.21 & 67.08 & 67.33 & 45.47 & 51.22 \\
ug & 19.37 & 52.32 & 26.88 & 28.23 & 52.12 & 63.89 \\
mt & 60.93 & 77.14 & 24.72 & 22.94 & 57.77 & 56.06 \\
\midrule
\textbf{Total avg.} & 31.93 & 60.26 & 45.18 & 45.12 & 58.83 & 68.68 \\
\bottomrule
\end{tabular}
\caption{Comparison of F1 Scores for LLaMA-3 Baseline (fine-tuned with a task adapter) and LLaMA-3+\texttt{Seq\_bn\_inv} on TC, NER, and SA. All results are averaged over 3 independent runs with different random seeds.}
\label{tab:llama3_adapter}
\end{table}

\end{document}